\documentclass[12pt,journal,draftcls,a4paper,onecolumn]{IEEEtran} 
\usepackage{filecontents,lipsum}
\usepackage[noadjust]{cite}
\usepackage{multicol,lipsum}
\usepackage{tikz,pgf} 
\usetikzlibrary{arrows,automata} 
\usepackage{verbatim}
\usetikzlibrary{positioning,shapes.geometric}
\usepackage{changes}
\usepackage{amsmath}
\usepackage{algpseudocode}
\usepackage{tabularx} 
\usepackage{array}
\usepackage{float}
\usepackage{amsmath}
\usepackage{amssymb}
\usepackage{multirow}
\usepackage{algorithm}
\usepackage{mathtools,xparse}
\DeclarePairedDelimiter{\norm}{\lVert}{\rVert}
\usepackage{amssymb}
\usepackage{array}
\usepackage[hidelinks]{hyperref}
\usepackage{tabularx} 
\usepackage{enumitem}
\usepackage{lipsum}
\usepackage{comment}
\usepackage{color}
\usepackage{graphicx}
\usepackage{subfig}
\usepackage{amsmath}
\usepackage{multirow}
\usepackage[all,poly]{xy}
\usepackage{layouts}
\usepackage{algpseudocode}
\usepackage{array}
\usepackage{fixltx2e}
\usepackage[font=small,labelfont=bf]{caption}
\usepackage[]{siunitx}
\usepackage{textcomp}
\newcolumntype{L}[1]{>{\raggedright\let\newline\\\arraybackslash\hspace{0pt}}m{#1}}
\newcolumntype{C}[1]{>{\centering\let\newline\\\arraybackslash\hspace{0pt}}m{#1}}
\newcolumntype{R}[1]{>{\raggedleft\let\newline\\\arraybackslash\hspace{0pt}}m{#1}}

\ifCLASSINFOpdf

\else

\fi

\begin{document}

\def\infinity{\rotatebox{90}{8}}

\def\bsa{{\boldsymbol{a}}}
\def\bsb{{\boldsymbol{b}}}
\def\bsc{{\boldsymbol{c}}}
\def\bsd{{\boldsymbol{d}}}
\def\bse{{\boldsymbol{e}}}
\def\bsf{{\boldsymbol{f}}}
\def\bsg{{\boldsymbol{g}}}
\def\bsh{{\boldsymbol{h}}}
\def\bsi{{\boldsymbol{i}}}
\def\bsj{{\boldsymbol{j}}}
\def\bsk{{\boldsymbol{k}}}
\def\bsl{{\boldsymbol{l}}}
\def\bsm{{\boldsymbol{m}}}
\def\bsn{{\boldsymbol{n}}}
\def\bso{{\boldsymbol{o}}}
\def\bsp{{\boldsymbol{p}}}
\def\bsq{{\boldsymbol{q}}}
\def\bsr{{\boldsymbol{r}}}
\def\bss{{\boldsymbol{s}}}
\def\bst{{\boldsymbol{t}}}
\def\bsu{{\boldsymbol{u}}}
\def\bsv{{\boldsymbol{v}}}
\def\bsw{{\boldsymbol{w}}}
\def\bsx{{\boldsymbol{x}}}
\def\bsy{{\boldsymbol{y}}}
\def\bsz{{\boldsymbol{z}}}

\def\bs0{{\boldsymbol{0}}}
\def\bflambda{{\boldsymbol{\lambda}}}
\def\bftheta{{\boldsymbol{\theta}}}
\def\bfphi{{\boldsymbol{\phi}}}
\def\bfSigma{{\boldsymbol{\Sigma}}}
\def\bfDelta{{\boldsymbol{\Delta}}}
\def\bfmu{{\boldsymbol{\mu}}}
\def\bfalpha{{\boldsymbol{\alpha}}}
\def\bfDelta{{\boldsymbol{\Delta}}}
\def\bfOmega{{\boldsymbol{\Omega}}}
\def\bfTheta{{\boldsymbol{\Theta}}}

\def\bsA{{\boldsymbol{A}}}
\def\bsB{{\boldsymbol{B}}}
\def\bsC{{\boldsymbol{C}}}
\def\bsD{{\boldsymbol{D}}}
\def\bsE{{\boldsymbol{E}}}
\def\bsF{{\boldsymbol{F}}}
\def\bsG{{\boldsymbol{G}}}
\def\bsH{{\boldsymbol{H}}}
\def\bsI{{\boldsymbol{I}}}
\def\bsJ{{\boldsymbol{J}}}
\def\bsK{{\boldsymbol{K}}}
\def\bsL{{\boldsymbol{L}}}
\def\bsM{{\boldsymbol{M}}}
\def\bsN{{\boldsymbol{N}}}
\def\bsO{{\boldsymbol{O}}}
\def\bsP{{\boldsymbol{P}}}
\def\bsQ{{\boldsymbol{Q}}}
\def\bsR{{\boldsymbol{R}}}
\def\bsS{{\boldsymbol{S}}}
\def\bsT{{\boldsymbol{T}}}
\def\bsU{{\boldsymbol{U}}}
\def\bsV{{\boldsymbol{V}}}
\def\bsW{{\boldsymbol{W}}}
\def\bsX{{\boldsymbol{X}}}
\def\bsY{{\boldsymbol{Y}}}
\def\bsZ{{\boldsymbol{Z}}}


\def\bfa{{\mathbf{a}}}
\def\bfb{{\mathbf{b}}}
\def\bfc{{\mathbf{c}}}
\def\bfd{{\mathbf{d}}}
\def\bfe{{\mathbf{e}}}
\def\bff{{\mathbf{f}}}
\def\bfg{{\mathbf{g}}}
\def\bfh{{\mathbf{h}}}
\def\bfi{{\mathbf{i}}}
\def\bfj{{\mathbf{j}}}
\def\bfk{{\mathbf{k}}}
\def\bfl{{\mathbf{l}}}
\def\bfm{{\mathbf{m}}}
\def\bfn{{\mathbf{n}}}
\def\bfo{{\mathbf{o}}}
\def\bfp{{\mathbf{p}}}
\def\bfq{{\mathbf{q}}}
\def\bfr{{\mathbf{r}}}
\def\bfs{{\mathbf{s}}}
\def\bft{{\mathbf{t}}}
\def\bfu{{\mathbf{u}}}
\def\bfv{{\mathbf{v}}}
\def\bfw{{\mathbf{w}}}
\def\bfx{{\mathbf{x}}}
\def\bfy{{\mathbf{y}}}
\def\bfz{{\mathbf{z}}}

\def\bfA{{\mathbf{A}}}
\def\bfB{{\mathbf{B}}}
\def\bfC{{\mathbf{C}}}
\def\bfD{{\mathbf{D}}}
\def\bfE{{\mathbf{E}}}
\def\bfF{{\mathbf{F}}}
\def\bfG{{\mathbf{G}}}
\def\bfH{{\mathbf{H}}}
\def\bfI{{\mathbf{I}}}
\def\bfJ{{\mathbf{J}}}
\def\bfK{{\mathbf{K}}}
\def\bfL{{\mathbf{L}}}
\def\bfM{{\mathbf{M}}}
\def\bfN{{\mathbf{N}}}
\def\bfO{{\mathbf{O}}}
\def\bfP{{\mathbf{P}}}
\def\bfQ{{\mathbf{Q}}}
\def\bfR{{\mathbf{R}}}
\def\bfS{{\mathbf{S}}}
\def\bfT{{\mathbf{T}}}
\def\bfU{{\mathbf{U}}}
\def\bfV{{\mathbf{V}}}
\def\bfW{{\mathbf{W}}}
\def\bfX{{\mathbf{X}}}
\def\bfY{{\mathbf{Y}}}
\def\bfZ{{\mathbf{Z}}}


\def\bbA{{\mathbb{A}}}
\def\bbB{{\mathbb{B}}}
\def\bbC{{\mathbb{C}}}
\def\bbD{{\mathbb{D}}}
\def\bbE{{\mathbb{E}}}
\def\bbF{{\mathbb{F}}}
\def\bbG{{\mathbb{G}}}
\def\bbH{{\mathbb{H}}}
\def\bbI{{\mathbb{I}}}
\def\bbJ{{\mathbb{J}}}
\def\bbK{{\mathbb{K}}}
\def\bbL{{\mathbb{L}}}
\def\bbM{{\mathbb{M}}}
\def\bbN{{\mathbb{N}}}
\def\bbO{{\mathbb{O}}}
\def\bbP{{\mathbb{P}}}
\def\bbQ{{\mathbb{Q}}}
\def\bbR{{\mathbb{R}}}
\def\bbS{{\mathbb{S}}}
\def\bbT{{\mathbb{T}}}
\def\bbU{{\mathbb{U}}}
\def\bbV{{\mathbb{V}}}
\def\bbW{{\mathbb{W}}}
\def\bbX{{\mathbb{X}}}
\def\bbY{{\mathbb{Y}}}
\def\bbZ{{\mathbb{Z}}}


\def\dsA{{\mathds{A}}}
\def\dsB{{\mathds{B}}}
\def\dsC{{\mathds{C}}}
\def\dsD{{\mathds{D}}}
\def\dsE{{\mathds{E}}}
\def\dsF{{\mathds{F}}}
\def\dsG{{\mathds{G}}}
\def\dsH{{\mathds{H}}}
\def\dsI{{\mathds{I}}}
\def\dsJ{{\mathds{J}}}
\def\dsK{{\mathds{K}}}
\def\dsL{{\mathds{L}}}
\def\dsM{{\mathds{M}}}
\def\dsN{{\mathds{N}}}
\def\dsO{{\mathds{O}}}
\def\dsP{{\mathds{P}}}
\def\dsQ{{\mathds{Q}}}
\def\dsR{{\mathds{R}}}
\def\dsS{{\mathds{S}}}
\def\dsT{{\mathds{T}}}
\def\dsU{{\mathds{U}}}
\def\dsV{{\mathds{V}}}
\def\dsW{{\mathds{W}}}
\def\dsX{{\mathds{X}}}
\def\dsY{{\mathds{Y}}}
\def\dsZ{{\mathds{Z}}}

\def\calh{{\mathcal{h}}}
\def\calU{{\mathcal{U}}}
\def\calu{{\mathcal{u}}}
\def\calS{{\mathcal{S}}}
\def\calV{{\mathcal{V}}}
\def\calv{{\mathcal{v}}}
\def\calP{{\mathcal{P}}}
\def\calA{{\mathcal{A}}}
\def\calB{{\mathcal{B}}}
\def\calC{{\mathcal{C}}}
\def\calD{{\mathcal{D}}}
\def\calE{{\mathcal{E}}}
\def\calF{{\mathcal{F}}}
\def\calG{{\mathcal{G}}}
\def\calH{{\mathcal{H}}}
\def\calI{{\mathcal{I}}}
\def\calJ{{\mathcal{J}}}
\def\calK{{\mathcal{K}}}
\def\calL{{\mathcal{L}}}
\def\calM{{\mathcal{M}}}
\def\calN{{\mathcal{N}}}
\def\calO{{\mathcal{O}}}
\def\calP{{\mathcal{P}}}
\def\calQ{{\mathcal{Q}}}
\def\calR{{\mathcal{R}}}
\def\calS{{\mathcal{S}}}
\def\calT{{\mathcal{T}}}
\def\calU{{\mathcal{U}}}
\def\calV{{\mathcal{V}}}
\def\calW{{\mathcal{W}}}
\def\calx{{\mathcal{x}}}
\def\calX{{\mathcal{X}}}
\def\caly{{\mathcal{y}}}
\def\calY{{\mathcal{Y}}}
\def\calZ{{\mathcal{Z}}}

\setlength{\belowcaptionskip}{-7pt}

\title{Deconvolution and Restoration of Optical Endomicroscopy Images}
\author{Ahmed~Karam Eldaly,~\IEEEmembership{Student~Member,~IEEE,} Yoann~Altmann,~\IEEEmembership{Member,~IEEE,}
        Antonios~Perperidis, Nikola Krstaji\'c, Tushar R. Choudhary, Kevin Dhaliwal,
      and~Stephen~McLaughlin,~\IEEEmembership{Fellow,~IEEE}
\thanks{A. K. Eldaly, Y. Altmann, A. Perperidis and S. McLaughlin are with the Institute of Sensors, Signals and Systems, School of Engineering and Physical Sciences, Heriot-Watt University, Edinburgh, UK. (Emails: \{AK577; Y.Altmann; A.Perperidis; S.Mclaughlin\}@hw.ac.uk)}
\thanks{T. R. Choudhary is with the Institute of Biological Chemistry, Biophysics and Bioengineering, Heriot-Watt University, Edinburgh, United Kingdom (Email: T.Choudhary@hw.ac.uk)}
\thanks{N. Krstaji\'c and K. Dhaliwal are with the EPSRC IRC Hub in Optical Molecular Sensing \& Imaging, MRC Centre for Inflammation Research, Queen's Medical Research Institute, University of Edinburgh, Edinburgh, UK (Emails: \{N.Krstajic; Kev.Dhaliwal\}@ed.ac.uk)}
\thanks{This work was supported by the EPSRC via grant EP/K03197X/1 and the Royal Academy of Engineering through the research fellowship scheme.}}

\maketitle

\begin{abstract}
Optical endomicroscopy (OEM) is an emerging technology platform with preclinical and clinical imaging applications. Pulmonary OEM via fibre bundles has the potential to provide \emph{in vivo, in situ} molecular signatures of disease such as infection and inflammation. However, enhancing the quality of data acquired by this technique for better visualization and subsequent analysis remains a challenging problem. Cross coupling between fiber cores and sparse sampling by imaging fiber bundles are the main reasons for image degradation, and poor detection performance (i.e., inflammation, bacteria, etc.). In this work, we address the problem of deconvolution and restoration of OEM data. We propose a hierarchical Bayesian model to solve this problem and compare three estimation algorithms to exploit the resulting joint posterior distribution. The first method is based on Markov chain Monte Carlo (MCMC) methods, however, it exhibits a relatively long computational time. The second and third algorithms deal with this issue and are based on a variational Bayes (VB) approach and an alternating direction method of multipliers (ADMM) algorithm respectively. Results on both synthetic and real datasets illustrate the effectiveness of the proposed methods for restoration of OEM images.
\end{abstract}


\IEEEpeerreviewmaketitle

\section{Introduction}
\IEEEPARstart{P}{neumonia} is a major cause of morbidity and mortality in mechanically ventilated patients in intensive care \cite{chastre2002ventilator}. However, the accurate diagnosis and monitoring of suspected pneumonia remain challenging \cite{johnston2010novel}. Current methodologies consist of culturing bronchoalveolar lavage fluid (BALF) retrieved from bronchoscopy, but this often takes 48 hours to yield a result which still has low specificity and sensitivity \cite{baselski1994bronchoscopic}. Structural imaging with X-ray or computed tomography (CT) scans are also often non-diagnostic. 

Optical endomicroscopy (OEM) is an emerging, optical fibre-based medical imaging modality with utility in a range of clinical indications and organ systems, including gastro-intestinal, urological and respiratory tracts. The technology employs a proximal light source, laser scanning or Light Emitting Diode (LED) illumination, linked to a flexible fibre bundle, performing microscopic fluorescent imaging at its distal end. The diameter of the packaged fibre can be $<\SI{500}{\micro\metre}$ , enabling the real-time imaging of tissues that were previously inaccessible through conventional endoscopy. Probe-based confocal laser endomicroscopy, is currently the most widely used clinical OEM platform approved for clinical use. However, there have recently been a number of studies describing novel, flexible, versatile and low-cost OEM architectures \cite{pierce2011high, shin2010fiber, hong2016smartphone}, employing wide-field LED illumination sources, capable of imaging at multiple acquisition wavelengths \cite{krstajic2016two}. Wide-field fiber optic imaging devices, such as the one being developed by our group provide sparse and usually irregularly-spaced intensity readings of the scene, due to the irregular packing of the fibre cores within the fibre bundle. Fibre bundles usually contain approximately 25,000 fibre cores that are transmitting and collecting the light simultaneously. Note that it is only the fibre cores which contain information while the cladding, (the space between the fibre cores), does not.

One of the main challenges of OEM images is enhancing the restoration of the signals at the receiver for better image visualization and/or subsequent analysis. Fiber core cross coupling is one of the main reasons for image degradation in this type of imaging \cite{wood2017quantitative, reichenbach2007numerical}. In confocal endomicroscopy, the detector pinhole can mask out light coupled to neighbouring cores before reaching the detector. Consequently, the effect of inter-core coupling in imaging capabilities is inherently of greater importance in wide-field endomicroscopy. Perperidis et al. \cite{perperidis2017characterization} have quantified the average spread of inter-core coupled light, with approximately a third of the overall light coupling to neighbouring cores. Consequently, cross coupling causes severe blurring in the resulting images, whose restoration is formulated as an inverse problem. We will discuss in detail cross coupling effects in Section \ref{sec:ProblemFormulation}. In this work, we consider a noisy observation vector $\bfy$, of an original intensity vector $\bfx$, that is modelled by the following linear forward model
\begin{equation}
\begin{aligned}
\bfy = \bfA\bfx + \bfw,
\label{eq:RestorationModel}
\end{aligned}
\end{equation}
where $\bfA$ is the matrix representing a linear operator which can model different degradation. Here, $\bfA$ models fiber core cross coupling and/or spatial blur. We specify the dimensions of the variables later in the text. In \eqref{eq:RestorationModel}, the vector $\bfw$ stands for additive noise, modelling observation noise and model mismatch and is assumed to be a white Gaussian noise sequence. In wide-field OEM, the constant background fluorescence of the fiber bundle \cite{udovich2008spectral, krstajic2016two}, is significant (between 90\% and 60\% of the total signal) providing a significant offset to all fluorescence measurements from tissue. Hence, the total noise level does not depend on the tissue signal level. Also, we consider applications where the photon flux is high ($> 500$ photoelectrons generated per pixel per typical exposure time 50 ms). Therefore, the Gaussian noise assumption holds \cite{sarder2006deconvolution, cannell2006image, qin2014shearlet}.

The problem of estimating $\bfx$ from $\bfy$ is an ill-posed linear inverse problem (LIP); i.e., the matrix $\bfA$ is singular or very ill-conditioned. Consequently, this problem requires additional regularization (or prior information, in Bayesian inference terms) in order to reduce uncertainties and improve estimation performance. State-of-the-art algorithms for solving such problems can be split into either convex optimization or Bayesian methods.

In \cite{vsroubek2003multichannel, figueiredo2009fast, afonso2011non, zhou2014fast}, the problem of estimating $\bfx$ given $\bfy$ is formulated as an unconstrained optimization problem as follows
\begin{equation} 
\begin{aligned}
& \underset{\bfx}{\text{minimize}}
& & \frac{1}{2}\norm*{\bfA\bfx - \bfy}^2_2  + \lambda\phi(\bfx) + i_{\mathbb{R}^+}(\bfx),
\label{eq:Restoration2}
\end{aligned}
\end{equation}
where $\phi(\cdot)$ is a regularization function, $\norm{.}_2$ is the standard $\ell_2$-norm, $\lambda \in \mathbb{R}_+$ is a regularization parameter, and $i_{\mathbb{R}^+}(\bfx)$ is the indicator function defined on the positive set of $\bfx$. For solving problems of the form \eqref{eq:Restoration2}, state-of-the-art algorithms potentially belonging to the iterative shrinkage/thresholding family \cite{vsroubek2003multichannel, figueiredo2009fast, afonso2011non, zhou2014fast} can be used. In \cite{afonso2010fast, figueiredo2009fast}, the unconstrained problem in Eq.\eqref{eq:Restoration2} is solved by an algorithm called split augmented Lagrangian shrinkage algorithm (SALSA) which is based on variable splitting \cite{courant1943variational, wang2008new}.

Alternatively, many studies have considered hierarchical Bayesian models to solve the deconvolution and restoration problem \cite{wong2015bayesian, bishop2008blind, ruiz2015variational, al2004statistical, al2011three, pankajakshan2009blind, babacan2009variational, babacan2011variational, molina2006blind, babacan2008parameter}. These models offer a flexible and consistent methodology to deal with uncertainty in inference when limited amount of data or information is available. Moreover, other unknown parameters can be jointly estimated within the algorithm such as noise variance(s) and regularization parameters. As such, they represent an attractive way to tackle ill-posed problems such as the one considered in this work. These methods rely on selecting an appropriate prior distribution for the unknown image and other unknown parameters. The full posterior distribution can then be derived from the Bayes' rule, and then exploited by optimization or simulation-based (Markov chain Monte Carlo) methods.

The main contributions of this work are fourfold:

\begin{enumerate}
\item We address the problem of deconvolution and restoration in OEM. To the best of our knowledge, it is the first time this problem is addressed in a statistical framework by using a hierarchical Bayesian model. 

\item We develop algorithms dedicated to irregularly sampled images which do not rely on strong assumptions about the spatial structure of the sampling patterns. The developed methods can thus be applied to a wide range of imaging systems, and fiber bundle designs.

\item We derive three estimation algorithms associated with the proposed hierarchical Bayesian model and compare them using extensive simulations conducted using controlled and real data. The first algorithm generates samples distributed according to the posterior distribution using Markov chain Monte Carlo (MCMC) methods \cite{robert2013monte}. This approach also allows the estimation of the hyperparameters associated with the priors. However, as mentioned previously, the resulting MCMC-based algorithm presents a high computational complexity. The second and third algorithms deal with this limitation and approximate the joint posterior distribution. The second algorithm uses the variational Bayes (VB) methodology \cite{mackay2003information, beal2003variational} to approximate the joint posterior distribution by minimizing the Kullback–Leibler (KL) divergence between the true posterior distribution and its approximation \cite{kullback1997information}. It can also estimate the hyperparameters associated with the prior distributions, and hence it is totally unsupervised, as is the MCMC-based method. The third algorithm is based on the alternating direction method of multipliers (ADMM). Although the low computation complexity of this algorithm, the hyperparameters associated with the priors need to be chosen carefully by the user, and hence it is considered as a semi-supervised method.

\item We use Gaussian Processes (GP) to interpolate the resulting samples to provide a meaningful image and quantify uncertainties at each interpolated sample. 
\end{enumerate}

The remaining sections of the paper are organized as follows. Section \ref{sec:ProblemFormulation} discusses the cross coupling problem and formulates the problem of deconvolution and restoration of OEM data. The proposed hierarchical Bayesian model is then presented in Section \ref{sec:BayesianModel}. Section \ref{sec:Estimation} introduces the three proposed estimation algorithms based on MCMC and optimization. Results of simulations conducted using synthetic and real datasets are discussed in Section \ref{sec:Results} and Section \ref{sec:RealData}, respectively. Conclusions and future work are finally reported in Section \ref{sec:Conclusion}.
\vspace*{-0.2cm}
\section{Problem Formulation}
\label{sec:ProblemFormulation}
\begin{figure}[!h]
	\centering
		\includegraphics[width=0.65\textwidth]{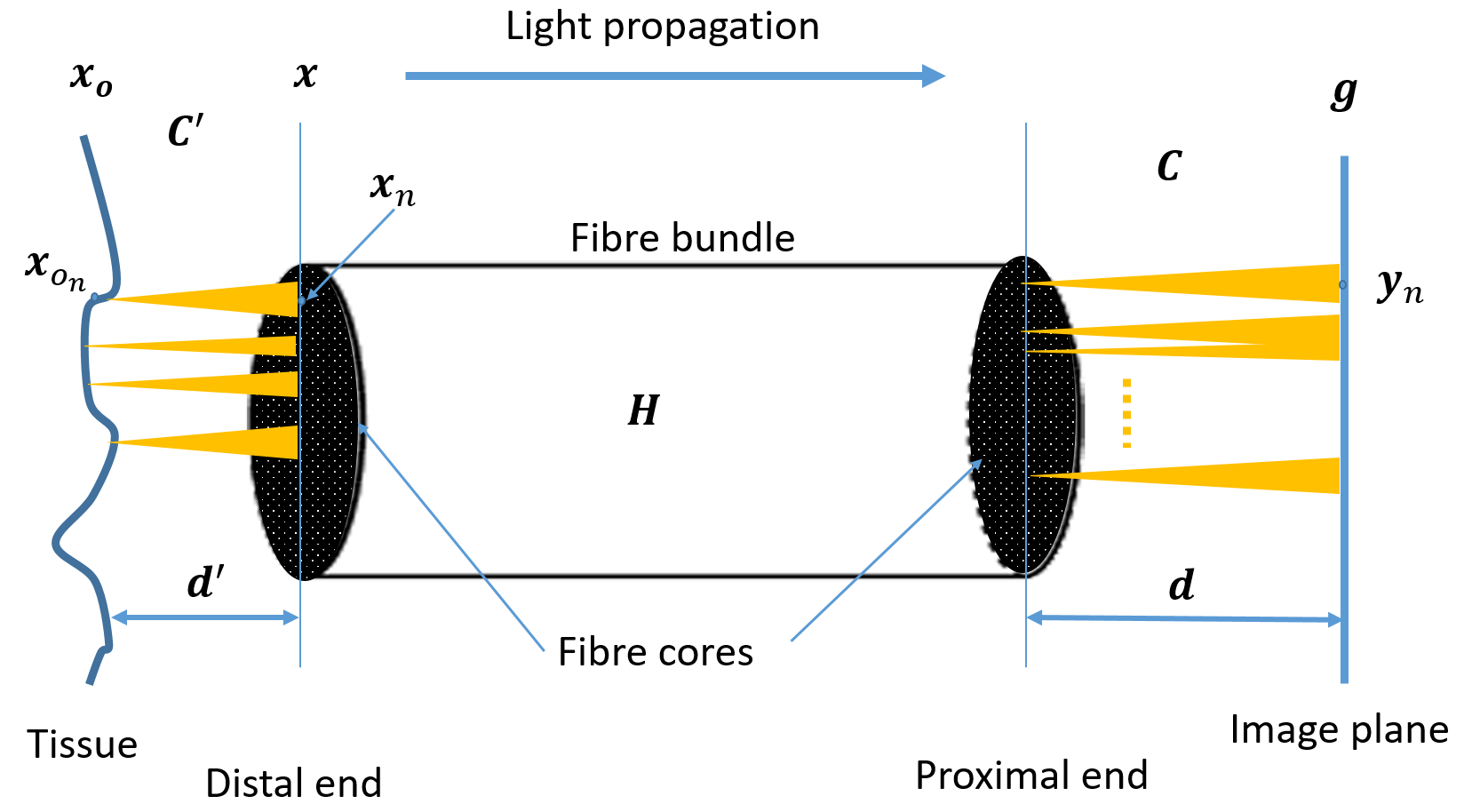}
	\caption{Schematic diagram showing the forward model in OEM.}
	\label{fig:BlurringEffect}
\end{figure}

Fig. \ref{fig:BlurringEffect} illustrates what happens in the fibre bundle when receiving fluorescent light from an object being imaged. The vectors $\bfx_o$, $\bfx$, and $\bfg$ represent light intensities at the object being imaged (tissue in this case), at the distal end of the fibre bundle, and at the image plane respectively. The transform $\bfH$ represents the cross coupling effect defined later in the text, $\bfC$ represents the spatial blur acting between the proximal end of the fibre bundle and the image plane, whereas $\bfC^\prime$ is that between the distal end of the fibre bundle and the tissue being imaged. The two spatial blurs $\bfC$ and $\bfC^\prime$ are spatially variant, $\bfC$ can be characterized as the distance $d$ between the image plane and the proximal end of the fibre is known, whereas $\bfC^\prime$ cannot be fully characterized as $d^\prime$ is unknown and the frames here are analyzed independently. Hence, to overcome this problem, we aim to recover the intensity vector $\bfx$ rather than $\bfx_o$.

\begin{figure}[h]
	\centering
		\includegraphics[width=0.5\textwidth]{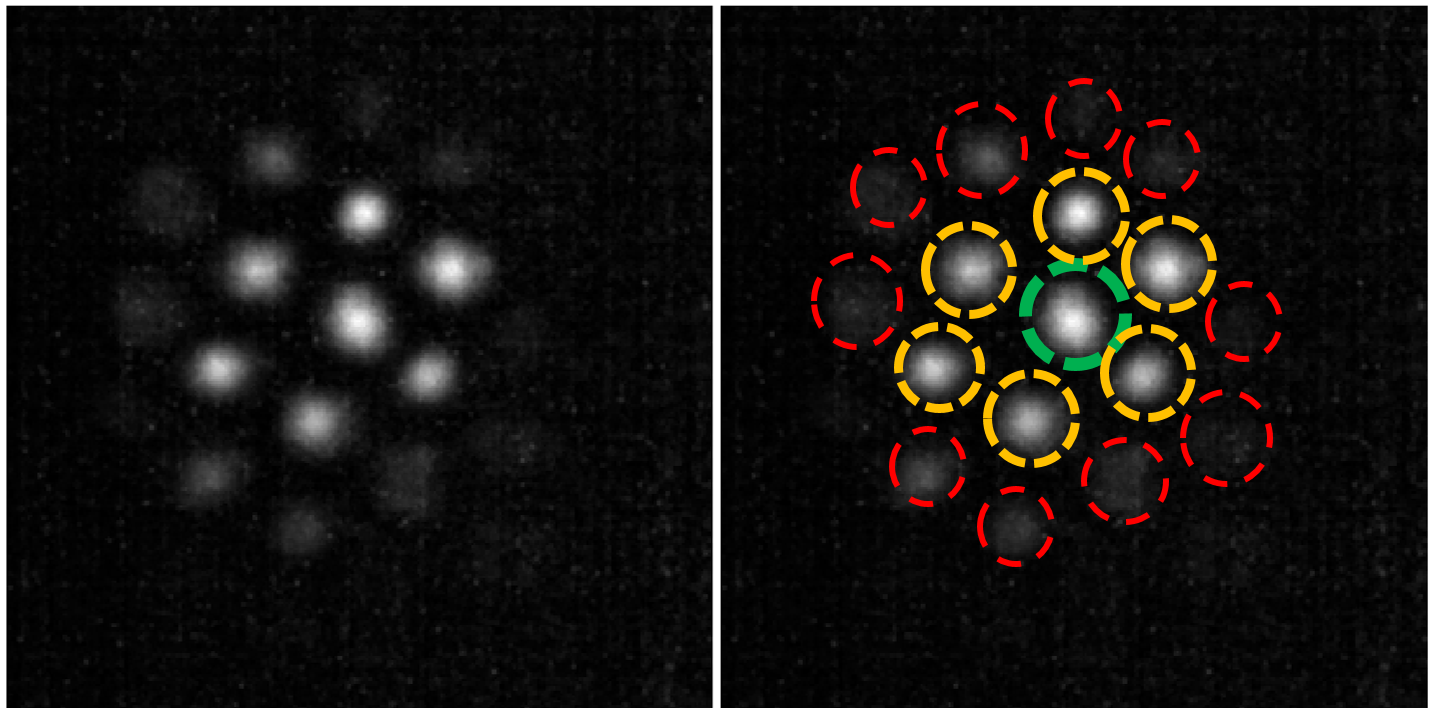}
	\caption{Example of cross coupling between fiber cores, the green circle represents the central illuminated core and the yellow and red ones represent the immediate and further neighbours respectively.}
	\label{fig:Cross_Coupling}
\end{figure}

Fig. \ref{fig:Cross_Coupling} provides and illustrative example of cross coupling between fiber cores. If an individual fiber core is illuminated in $\bfx$, the neighbouring cores in $\bfg$ will be affected by a specific percentage of the incident light on the illuminated core. Experimental results in current fiber bundle (which might be different for other bundles) showed that around 61\% of the light transmitted through a single core remains in that core, around 34\% migrates to the immediate neighbouring cores, around 4\% to the second order neighbours and less than 1\% to the third, fourth, and fifth order neighbours \cite{perperidis2017characterization}.

\begin{figure}[h]
	\centering
		\includegraphics[width=0.81\textwidth]{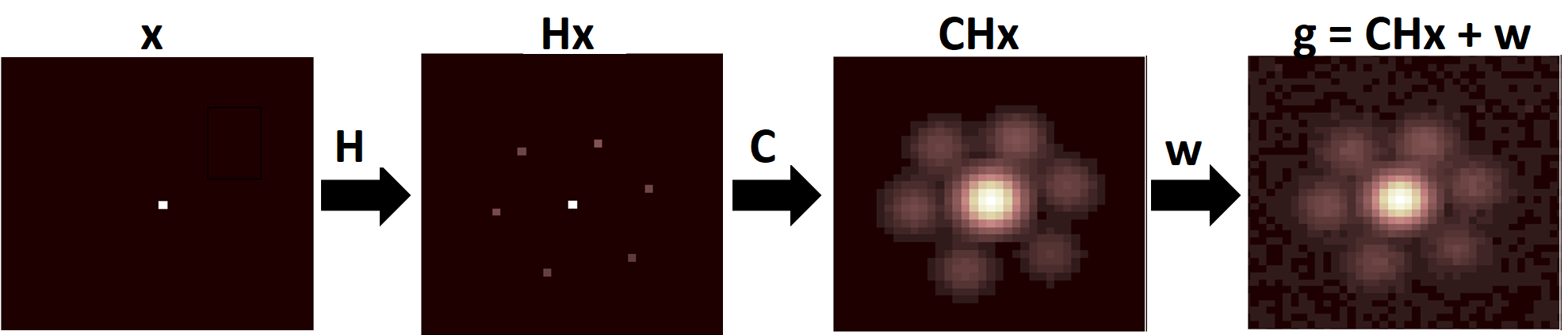}
	\caption{Representation of the endomicroscopy system output images.}
	\label{fig:Graphical_Model}
\end{figure}

\vspace*{+0.3cm}
Fig. \ref{fig:Graphical_Model} illustrates how we construct the forward observation model to mimic the same output as the endomicroscopy imaging system. The first image on the left-hand side of the figure represents the illumination of one fiber core. This results in cross coupling to the neighbouring cores (convolution with a first linear operator $\bfH$), then the spatial blurring effect around each fiber core (convolution with a second linear operator $\bfC$) and finally the fourth image of the figure shows the final system output after adding white Gaussian noise. 

\vspace*{+0.3cm}
The linear model in \eqref{eq:RestorationModel} can now be written as 
\begin{equation}
\begin{aligned}
\bfg = \bfC\bfH\bfx + \bfw,
\label{eq:Formulation_1}
\end{aligned}
\end{equation}
where $\bfA$ in \eqref{eq:RestorationModel} is replaced by $\bfC\bfH$ in \eqref{eq:Formulation_1}, the vector $\bfg$ is the observed data matrix, and $\bfx$ is the image to be restored.

From preliminary results, we propose to model cross-coupling by an isotropic zero mean 2D generalized Gaussian kernel applied to the fiber intensities \cite{perperidis2017characterization} as follows

\begin{equation}
\begin{aligned}
[\bfH]_{i,j} = \exp\left(-\left(\frac{d_{i,j}}{\alpha_{\bfH}}\right)^{\beta_\bfH}\right),
\label{eq:CrossCoupling_Form}
\end{aligned}
\end{equation}
where $d_{i, j}$ denotes the euclidean distance between the cores (or spatial locations) $i$ and $j$, which corresponds to approximately 3.3 pixels between neighbouring cores. From \eqref{eq:CrossCoupling_Form}, it can be seen that neighbouring fiber cores will be more closely coupled than distant ones. The values of $\alpha_\bsH$ and $\beta_\bsH$, which control the amount of cross-coupling (the higher, the more coupling) and which are system dependent, are adjusted from preliminary measurements (calibration). Note that other cross-coupling models could also be considered instead of \eqref{eq:CrossCoupling_Form} depending on the imaging system used. 

\begin{figure}
	\centering
		\includegraphics[width=0.6\textwidth]{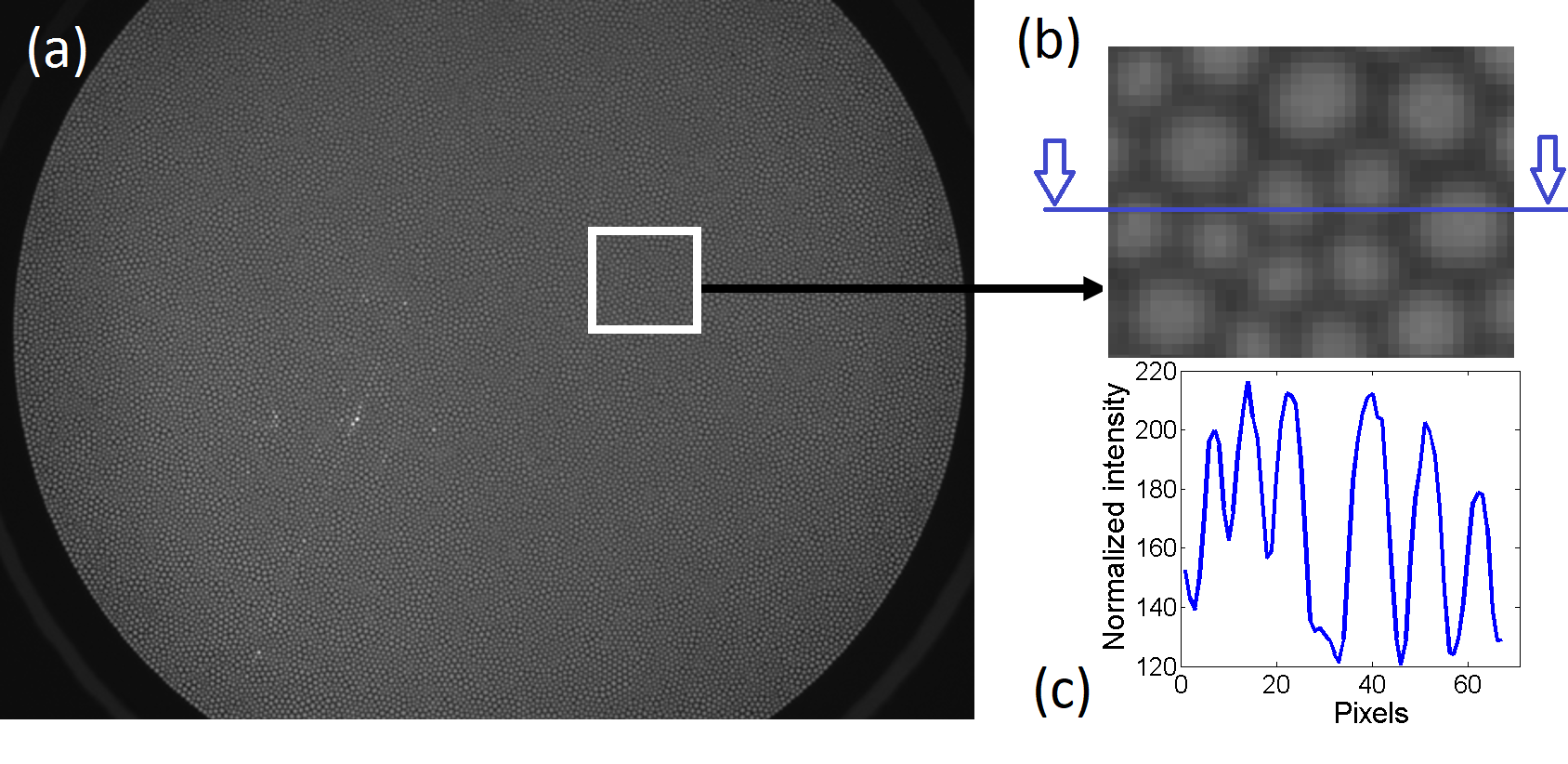}
	\caption{(a) A background image, (b) a zoomed part of the image, and (c) the intensity profile across one line in the image.}
	\label{fig:Intensity_Profile}
\end{figure}

The spatial blur affecting each fiber core can be modelled by a Gaussian spatial filter, as illustrated in Fig. \ref{fig:Intensity_Profile}, which shows a background image i.e., an image from a sample presenting constant intensity, using an endomicroscopy imaging system, and a zoomed-in region of this image, bright and dark areas represent fiber cores and their cladding, respectively. The intensity profile across one line in this image is a series of Gaussian kernels. However, the variation of the shape and width of the kernels is due to the variation in core sizes.

Due to the variation in core sizes, the blurring kernel $\bfC$ varies accordingly, and hence the cores tend to overlap. So the complete model in \eqref{eq:Formulation_1} becomes more complex, and potentially computationally expensive for long image sequences (videos). Indeed, there is no structure in $\bfC$ which allows us to compute $\bfC\bfH\bfx$ rapidly. Hence we propose a simplification of this model and represent each core by a single intensity value. The mean intensities of fibre core pixels could be used, but the overlap between the cores makes its computation difficult. Since the variation of the width of this blur is not too significant, the maximum intensity of each core is considered instead ($\bfy_n$ in Fig. \ref{fig:BlurringEffect}).

Following the above mentioned points, the model in \eqref{eq:Formulation_1} can be simplified to
\begin{equation} 
\begin{aligned}
& \bfy = \bfH\bfx + \bfw.
\label{eq:Bayesian_1}
\end{aligned}
\end{equation}

Assume that $N$ is the total number of pixels in the image, and $N_1$ representing number of fibre cores in the image, the input $\bfy \approx \bfC^+\bfg \in \mathbb{R}^{N_1}$, where $\bfC^+$ is the pseudo-inverse of $\bfC$, and the output $\bfx \in \mathbb{R}^{N_1}$ are two vectors representing central core intensities, where, $N_1 << N$, and $\bfH \in \mathbb{R}^{N_1 \times N_1}$. The noise $\bfw \in \mathbb{R}^{N_1}$ is assumed to be additive white noise which is independent and identically distributed (i.i.d) zero mean Gaussian noise with variance $\sigma^2$, denoted as $\bfw \sim \mathcal{N}(\bf0, \sigma^2\bfI)$, where $\sim$ means ``is distributed according to'' and $\bfI$ is the identity matrix.

The problem investigated in this paper is to estimate the actual intensity values $\bfx$, and the noise variance $\sigma^2$ from the observation vector $\bfy$. As mentioned previously, to solve this problem, we propose a hierarchical Bayesian model and a set of different estimation methods to estimate the unknown parameters.

\section{Hierarchical Bayesian Model}
\label{sec:BayesianModel}
This section introduces a hierarchical Bayesian model proposed to estimate the unknown parameter vector $\bfx$ and $\sigma^2$. This model is based on the likelihood function of the observations and on prior distributions assigned to the unknown parameters.
\subsection{Likelihood}
Eq. \eqref{eq:Bayesian_1} yields that $\bfy|(\bfx, \sigma^2) \sim \mathcal{N}(\bfH\bfx, \sigma^2\bfI)$. Consequently, the likelihood can be expressed as
\begin{equation}
\begin{aligned}
& f(\bfy |  \bfx, \sigma^2) = \left(\frac{1}{2\pi \sigma^2}\right)^{N_1/2} \exp\left(-\frac{\norm{\bfy - \bfH\bfx}^2_2}{2\sigma^2}\right).
\end{aligned}
\end{equation}

\subsection{Parameter Priors}
\subsubsection{Prior for the underlying intensity field $\bfx$}
\label{subsubsec:Xprior}
A truncated multivariate Gaussian distribution (MVG) is assigned to the intensity field $\bfx$.
\begin{equation}
\begin{aligned}
& f(\bfx | \gamma^2) \propto {\left(\gamma^2\right)}^{-d/2} \exp\left(- \frac{\bfx^T\bfDelta^{-1}\bfx}{2\gamma^2}\right) 1_{\mathbb{R}^+}(\bfx),
\label{eq:Bayesian_3}
\end{aligned}
\end{equation}
where $1_{\mathbb{R}^+}(\bfx)$ is the indicator function defined on the positive set of $\bfx$, $\gamma^2$ controls the global correlation between intensities, and the covariance matrix $\bfDelta$ which defines the spatial correlation between the cores is defined by
\begin{equation}
\begin{aligned}
[\bfDelta]_{n, n'} = \exp\left(-\left(\frac{d_{n, n'}}{\ell}\right)^\kappa\right),
\label{eq:CovarianceMRFandGP}
\end{aligned}
\end{equation}
where $d_{n, n'}$ denotes the distance between the spatial locations $n \text{ and } n'$, and $d=N_1$. Equations \eqref{eq:Bayesian_3} and \eqref{eq:CovarianceMRFandGP} promote smooth intensity variations between neighbours while ensuring that the prior dependence between neighbouring cores decrease as $d_{n, n'}$ increases. In this work $d_{n, n'}$ is the standard euclidean distance. The parameters $\ell, \kappa$ were learned from the irregular sampling pattern of the OEM system. Precisely, we used known images and selected $(\ell, \kappa)$ by maximum likelihood estimation, which occurs when $p(\ell, \kappa | \bfx)$ is at its greatest, which corresponds to maximizing $\log p(\ell, \kappa | \bfx)$.  While $\gamma^2$ is left unknown for each image, $(\ell, \kappa)$ are fixed in the rest of the simulations as the average values obtained with the training images.

Considering such a prior is equivalent to assuming a Gaussian process on $\bfx$, this allows us to interpolate the resulting deconvolved intensities using Gaussian processes \cite{rasmussen2006gaussian} as we will see in section \ref{sec:GPR}.

\subsubsection{Prior for the noise variance $\sigma^2$}
A conjugate inverse-Gamma $\mathcal{IG}$ prior is assigned to the noise variance $\sigma^2$  
\begin{equation}
\begin{aligned}
& f(\sigma^2 | \alpha, \beta) \sim \mathcal{IG}(\alpha, \beta),
\label{eq:Bayesian_6}
\end{aligned}
\end{equation}
where $\alpha = 10$ is fixed arbitrarily, while the hyperparameter $\beta$ is estimated within the algorithm.
\subsubsection{Prior for the hyperparameter $\beta$}
The hyperparameter associated with the parameter prior defined above is assigned to a conjugate Gamma distribution:
\begin{equation}
\begin{aligned}
& \beta \sim \mathcal{G}(\alpha_o, \beta_o),
\label{eq:Bayesian_5}
\end{aligned}
\end{equation}
where $\alpha_o$ and $\beta_o$ are fixed and user-defined parameters which might depend on the quality of the data to be recovered. In this work, we fixed $(\alpha_o, \beta_o) = (10, 0.1)$ arbitrarily.
\subsubsection{Prior for the hyperparameter $\gamma^2$}
To reflect the lack of prior knowledge about the regularization parameter $\gamma^2$ in \eqref{eq:Bayesian_3}, the following weakly informative conjugate inverse-Gamma prior is assigned to it.
\begin{equation}
\begin{aligned}
& \gamma^2 \sim \mathcal{IG}(\eta, \nu),
\label{eq:Bayesian_4}
\end{aligned}
\end{equation}
where $(\eta,\nu)$ are fixed to $(\eta,\nu)=(10^{-3},10^{-3})$. Note that we did not observe significance change in the results when changing these hyperparameters.\\

The next section derives the joint posterior distribution of the unknown parameters associated with the proposed Bayesian model.
\vspace*{-0.3cm}
\subsection{Joint posterior distribution}
Assuming the parameters $\bfx$ and $\sigma^2$ are \textit{a priori} independent, the joint posterior distribution of the parameter vector $\bfOmega = \{\bfx, \sigma^2\}$ and
hyperparameters $\bfphi = \{\beta, \gamma^2\}$ can be expressed as 
\begin{equation}
\begin{aligned}
& f(\bfOmega, \bfphi | \bfy) \propto f(\bfy | \bfOmega) f(\bfOmega | \phi) f(\bfphi),
\label{eq:Posterior_1}
\end{aligned}
\end{equation}
where
\begin{equation}
\begin{aligned}
& f(\bfOmega | \bfphi) = f(\bfx | \gamma^2) f(\sigma^2 | \beta), \text{ and } f(\bfphi) = f(\gamma^2) f(\beta).
\label{eq:Posterior_2}
\end{aligned}
\end{equation}

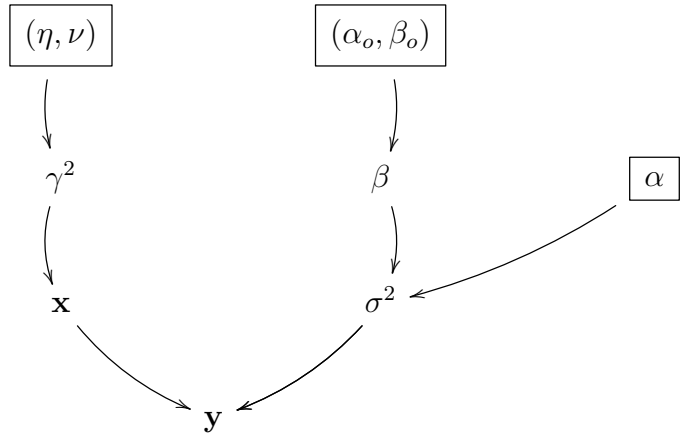
\begin{figure}
\centerline{ \xymatrix{
 *+<0.008in>+[F-]+{(\eta,\nu)} \ar@/_/[d] &  &   *+<0.008in>+[F-]+{(\alpha_o,\beta_o)} \ar@/^/[d] \\
 \gamma^2 \ar@/_/[d]   & &  \beta  \ar@/^/[d] & & *+<0.008in>+[F-]+{\alpha}  \ar@/^/[lld]\\
  \bfx \ar@/_/[rd]   &  & \sigma^2 \ar@/^/[ld] \ar@/^/[ld]  \\
   & \bfy & }} 
   \caption{Graphical model for the proposed hierarchical Bayesian model (fixed quantities appear in boxes).} \label{fig:DAG}
\end{figure}

The directed acyclic graph (DAG) summarizing the structure of proposed Bayesian model is depicted in Fig. \ref{fig:DAG}. This posterior distribution will be used to evaluate Bayesian estimators of $\bfTheta = \{\bfOmega, \bfphi\}$. For this purpose, we propose three algorithms: an MCMC-based approach and two optimization-based approaches, in which VB and ADMM are considered. The first approach uses an MCMC method to evaluate the minimum-mean-square-error (MMSE) estimator of $\bfTheta$ by generating samples according to the joint posterior distribution. Moreover, it allows the estimation of the hyperparameter vector $\bfphi$ along with the noise variance $\sigma^2$. However, it exhibits a relatively long computational time. The second and third algorithms which deal with this issue and provide fast MMSE estimate for the VB approach and MAP estimate for the ADMM approach. The VB approach approximates the joint posterior distribution in \eqref{eq:Posterior_1} by minimizing the Kullback-Leibler (KL) divergence between the true posterior distribution and its approximation \cite{kullback1997information}. The ADMM approach is achieved by maximizing the posterior distribution \eqref{eq:Posterior_1} with respect to (w.r.t.) $\bfTheta$. Note however, that the hyperparameters $\bfphi$ as well as $\sigma^2$ are fixed for this approach. The three estimation algorithms are described in the next section. 

\section{Bayesian Inference}
\label{sec:Estimation}
\subsection{MCMC algorithm}
To overcome the challenging derivation of Bayesian estimators associated with $f(\bfTheta | \bfy)$, we propose to use an efficient MCMC method to generate samples asymptotically distributed according to the posterior presented in \eqref{eq:Posterior_1}. More precisely, we consider a Gibbs sampler described next. The principle of the Gibbs sampler is to sample according to the conditional distributions of the posterior of interest [\cite{robert2013monte}, Chap. 10]. In this work, we propose to sample sequentially the elements of $\bfTheta$ using updates that are detailed below.

\subsubsection{Sampling the intensity field $\bfx$}
From \eqref{eq:Posterior_1}, since the prior \eqref{eq:Bayesian_3} is conjugate to the Gaussian distribution, the full conditional distribution of $\bfx$ is given by
\begin{equation}
\begin{aligned}
& f(\bfx | \bfy, \sigma^2) \sim \mathcal{N}_{\mathbb{R}^+}(\bfx; \bfmu, \bfSigma),
\label{eq:SampleX}
\end{aligned}
\end{equation}
where
\begin{equation}
\begin{aligned}
& \bfmu = \sigma^{-2}\bfSigma^T\bfH^T\bfy,\\
& \bfSigma = \left(\sigma^{-2}\bfH^T\bfH + \gamma^{-2}\bfDelta^{-1}\right)^{-1}.
\end{aligned}
\label{eq:IntensityPrior}
\end{equation}

Sampling from \eqref{eq:SampleX} can be achieved efficiently by using the Hamiltonian method proposed in \cite{pakman2014exact}.

\subsubsection{Sampling the noise variance $\sigma^2$}
By cancelling out the terms that don't depend on $\sigma^2$ from the posterior distribution in \eqref{eq:Posterior_1}, its conditional distribution can be written as
\begin{equation}
\begin{aligned}
& f(\sigma^2 | \bfy, \bfx) \sim \mathcal{IG}\left(\alpha + \frac{N_1}{2}, \beta + \frac{\norm{\bfy - \bfH\bfx}^2_2}{2}\right),
\label{eq:SampleSigma2}
\end{aligned}
\end{equation}
which is easy to sample from.
\subsubsection{Sampling the hyperparameters $\beta$ and $\gamma^2$}
It can be easily shown that $\beta$ can be sampled from the following Gamma distribution
\begin{equation}
\begin{aligned}
& f(\beta | \sigma^2) \sim \mathcal{G}\left(\alpha + \alpha_o, \frac{\sigma^2 \beta_o}{\sigma^2 + \beta_o}\right).
\label{eq:SampleBeta}
\end{aligned}
\end{equation}
In a similar fashion to the noise variance, $\gamma^2$ can be sampled from the following inverse-Gamma distribution
\begin{equation}
\begin{aligned}
& f(\gamma^2 | \bfx) \sim \mathcal{IG}\left(\eta + \frac{N_1}{2}, \nu + \frac{\bfx^T\bfDelta^{-1}\bfx}{2}\right).
\label{eq:SampleGamma2}
\end{aligned}
\end{equation}

The algorithm for generating samples asymptotically distributed according to the posterior distribution using Gibbs sampler is shown in Algorithm \autoref{Gibbs}.

\begin{algorithm}
\caption{Deconvolution via MCMC: Gibbs Sampling Algorithm}
\label{Gibbs}
\begin{algorithmic}[1]
\State \textbf{Fixed input parameters}: Number of burn-in iterations $N_{\text{bi}}$, total number of iterations $N_{\text{MC}}$
\State \textbf{Initializations} ($k = 0$)
\begin{itemize}
\item Set $\bfx^{(0)}$, ${\sigma^2}^{(0)}$, $\beta^{(0)}$, ${\gamma^2}^{(0)}$
\end{itemize}
\State \textbf{Repeat ($1 \leq k \leq N_{\text{MC}}$)}
\begin{itemize}
\item Sample $\bfx^{(k)}$ from \eqref{eq:SampleX} 
\item Sample ${\sigma^2}^{(k)}$ from \eqref{eq:SampleSigma2} 
\item Sample $\beta^{(k)}$ \ from \eqref{eq:SampleBeta} 
\item Sample ${\gamma^2}^{(k)}$ from \eqref{eq:SampleGamma2} 
\end{itemize}
\State \textbf{Set} $k = k + 1$.
\end{algorithmic}
\end{algorithm}

The posterior distribution mean or minimum mean square error (MMSE) estimator of $\bfx$ can be approximated by
\begin{equation}
\begin{aligned}
& \hat{\bfx} = \frac{1}{N_{\text{MC}} - N_{\text{bi}}}\sum_{t = N_{\text{bi} + 1}}^{N_{\text{MC}}} \bfx^{(t)},
\end{aligned}
\end{equation}
where the samples from the first $N_{\text{bi}}$ iterations (corresponding to the transient regime or burn-in period, which is determined visually from preliminary runs) of the sampler are discarded.
\subsection{Variational Bayes algorithm}
\label{subsec:VB}
For this approach, we consider an approximation of $p(\bfTheta| \bfy)$ by a simpler tractable distribution $q(\bfTheta)$ following the variational methodology \cite{beal2003variational}, moreover, here, we relax the positivity constraints about the intensity field vector $\bfx$. Note, however that the positivity constraints can be incorporated but the covariance matrix of the intensity field $\bfx$ would become more complicated \cite{vsmidl2006variational}, chap. 5. As will be shown in Sections \ref{sec:Results} and \ref{sec:RealData}, this constraint relaxation yields a fast estimation procedure providing estimation results which compete with the methods incorporating this constraint. The distribution $q(\bfTheta)$ will be found by minimizing the Kullback-Leibler (KL) divergence, between the actual posterior distribution and its approximation, given by \cite{kullback1997information} \cite{kullback1951information}

\begin{equation}
\begin{aligned}
\bfD_{\text{KL}}\left(q(\bfTheta) || p(\bfTheta| \bfy)\right) = \int q(\bfTheta) \log\left(\frac{q(\bfTheta)}{p(\bfTheta|\bfy)}\right) d\bfTheta,
\label{eq:KL}
\end{aligned}
\end{equation}
which is always non-negative and equal to zero only when $q(\bfTheta) = p(\bfTheta|\bfy)$. In order to obtain a tractable approximation, the family of distributions $q(\bfTheta)$ are restricted utilizing the mean field approximation \cite{parisi1988statistical} so that $q(\bfTheta) = q(\bfphi) q(\bfx) q(\sigma^2)$, where $q(\bfphi) = q(\gamma^2) q(\beta)$.

The lower bound of the KL divergence is given by
\begin{equation}
\begin{aligned}
p(\bfTheta, \bfy) \geq p(\bfy|\bfTheta)p(\bfTheta|\bfphi)p(\bfphi) = F(\bfTheta, \bfy).
\end{aligned}
\end{equation}

For $\mathcal{H}\in \{x, \sigma^2, \gamma^2, \beta\}$, let us denote by $\bfTheta_{\backslash \mathcal{H}}$, the subset of $\bfTheta$ with $\mathcal{H}$ removed; for instance, if $\mathcal{H} = \bfx$, $\bfTheta_{\backslash \bfx} = \{\sigma^2, \gamma^2, \beta\}$. Then utilizing the lower bound $\bfF(\bfTheta, \bfy)$ for the joint probability distribution in \eqref{eq:KL} we obtain an upper bound for the KL divergence as follows

\begin{equation}
\begin{aligned}
&\mathcal{M}\left(q(\bfTheta)\right) = \int q(\bfTheta) \log\left(\frac{q(\bfTheta)}{p(\bfTheta|\bfy)}\right) d\bfTheta \\
&\hspace{0.1cm}\leq \int q({\mathcal{H}})\left( \int q(\bfTheta_{\backslash \mathcal{H}}) \log\left(\frac{q(\mathcal{H})q(\bfTheta_{\backslash \mathcal{H}})}{F(\bfTheta,\bfy)}\right) d\bfTheta_{\backslash \mathcal{H}}\right)d\mathcal{H}\\
&\hspace{0.1cm}= \mathcal{M}\left(q(\mathcal{H})\right).
\label{eq:KLApprox}
\end{aligned}
\end{equation}
Therefore, we minimize this upper bound instead of minimizing the KL divergence in \eqref{eq:KL}. Note that the form of the inequality in \eqref{eq:KLApprox} suggests an alternating (cyclic) optimization strategy where the algorithm cycles through the unknown distributions and replaces each variable with a revised estimate given by the minimum of \eqref{eq:KLApprox} with the other distributions held constant. Thus, given $q(\bfTheta_{\backslash \mathcal{H}})$, the posterior distribution approximation $q({\mathcal{H}})$ can be computed by solving

\begin{equation}
\begin{aligned}
\hat{q}(\mathcal{H}) = \underset{q(\mathcal{H})}{\text{minimize}}
& & \bfD_{\text{KL}}\left( q(\bfTheta_{\backslash \mathcal{H}})q(\mathcal{H}) || F(\bfTheta, \bfy)\right).
\label{eq:KLApprox2}
\end{aligned}
\end{equation}

In order to solve this equation, we note that differentiating the integral on the right hand side in \eqref{eq:KLApprox} w.r.t. $q(\mathcal{H})$ results in (see \cite{miskin2000ensemble}, Eq. (2.28))

\begin{equation}
\begin{aligned}
\hat{q}(\mathcal{H}) = const \times \exp\left(E_{q(\bfTheta_{\backslash \mathcal{H}})}[\log F(\bfTheta, \bfy)]\right),
\label{eq:KLApprox3}
\end{aligned}
\end{equation}
where
\begin{equation}
\begin{aligned}
E{q(\bfTheta_{\backslash \mathcal{H}})}[\log F(\bfTheta, \bfy)] = \int \log F(\bfTheta, \bfy)q(\bfTheta_{\backslash \mathcal{H}})d\bfTheta_{\backslash \mathcal{H}}.
\end{aligned}
\end{equation}

We obtain the following iterative procedure to find $q(\bfTheta)$ by applying this minimization to each unknown in an alternating way

\begin{algorithm}
\caption{VB algorithm}
\label{alg:VB}
\begin{algorithmic}[1]
\State \textbf{Set} $k = 1$, \textbf{choose} $q^1(\sigma^2), q^1(\beta)$ and $q^1(\gamma^2)$, initial estimates of the distributions $q(\sigma^2), q(\beta)$ and $q(\gamma^2)$,
\State \textbf{repeat} (k = k + 1)

\State $q^k(\bfx) = \underset{q(\bfx)}{\text{minimize}} \int\int q^k(\bfTheta_{\backslash \bfx}) q(\bfx) \times \log\left(\frac{q^k(\bfTheta_{\backslash\bfx})q(\bfx)}{F(\bfTheta_{\backslash\bfx}^k, \bfx, \bfy)}\right) d\bfTheta_{\backslash\bfx} d\bfx$
\State $q^k(\sigma^2) = \underset{q(\sigma^2)}{\text{minimize}} \int\int q^k(\bfTheta_{\backslash\sigma^2}) q(\sigma^2) \times \log\left(\frac{q^k(\bfTheta_{\backslash\sigma^2})q(\sigma^2)}{F(\bfTheta_{\backslash\sigma^2}^k, \bfx, \bfy)}\right) d\bfTheta_{\backslash\sigma^2} d\sigma^2$
\State $q^k(\gamma^2) = \underset{q(\gamma^2)}{\text{minimize}} \int\int q^k(\bfTheta_{\backslash\gamma^2}) q(\gamma^2) \times \log\left(\frac{q^k(\bfTheta_{\backslash\gamma^2})q(\gamma^2)}{F(\bfTheta_{\backslash\gamma^2}^k, \bfx, \bfy)}\right) d\bfTheta_{\backslash\gamma^2} d\gamma^2$
\State $q^k(\beta) = \underset{q(\beta)}{\text{minimize}} \int\int q^k(\bfTheta_{\backslash\beta}) q(\beta) \times \log\left(\frac{q^k(\bfTheta_{\backslash\beta})q(\beta)}{F(\bfTheta_{\backslash\beta}^k, \beta, \bfy)}\right) d\bfTheta_{\backslash\beta} d\beta$
\State \textbf{until} some stopping criterion is satisfied.
\end{algorithmic}
\end{algorithm}

Now we detail the solutions at each step of algorithm \eqref{alg:VB} explicitly. 

\subsubsection{Updating intensity field vector $\bfx$}
From \eqref{eq:KLApprox3}, it can be shown that $q^k(\bfx)$ is an $N_1$-dimensional Gaussian distribution, rewritten as

\begin{equation}
\begin{aligned}
q^k(\bfx) = \mathcal{N}\left(\bfx; E_{q^k(\bfx)}(\bfx), \bfSigma_{q^k(\bfx)}(\bfx)\right),
\end{aligned}
\end{equation}
where the mean $E_{q^k(\bfx)}(\bfx)$ and covariance $\bfSigma_{q^k(\bfx)}(\bfx)$ of this normal distribution can be calculated from step 3 in Algorithm \ref{alg:VB} as 

\begin{subequations}\label{eq:IntensityFieldVB}
\begin{align}
& E_{q^k(\bfx)}(\bfx) = \frac{(\bfSigma_{q^k(\bfx)}(\bfx))^T\bfH^T\bfy}{E_{q^k(\sigma^2)}(\sigma^2)}, \\
& \bfSigma_{q^k(\bfx)}(\bfx) = \left(\frac{\bfH^T\bfH}{E_{q^k(\sigma^2)}(\sigma^2)} + \frac{\bfDelta^{-1}}{E_{q^k(\gamma^2)}(\gamma^2)}\right)^{-1}. 
\end{align}
\end{subequations}
\subsubsection{Updating noise variance $\sigma^2$}
It is easy to show from \eqref{eq:KLApprox3} that the noise variance follows an inverse-Gamma distribution given by

\begin{equation}
\begin{aligned}
q^k(\sigma^2) = \mathcal{IG}\left(\sigma^2; \frac{N_1}{2} + \alpha, E_{q^k(\beta)}(\beta) + E_{q^k(\bfx)}\left[ \norm{\bfy - \bfH\bfx}^2_2\right]\right),
\end{aligned}
\end{equation}
whose mean is given by
\begin{equation}
\begin{aligned}
E_{q^k(\sigma^2)}(\sigma^2) = \frac{E_{q^k(\beta)}(\beta) + E_{q^k(\bfx)}\left[ \norm{\bfy - \bfH\bfx}^2_2\right]}{N_1/2 + \alpha - 1},
\end{aligned}
\end{equation}
where
\begin{equation}
\begin{aligned}
E_{q^k(\bfx)}\left[ \norm{\bfy - \bfH\bfx}^2_2\right] = \norm{\bfy - \bfH E_{q^k(\bfx)}(\bfx)}^2_2 \\
\hspace{3cm} + \text{tr}\left(\bfH^T\bfH \bfSigma_{q^k(\bfx)}(\bfx)\right).
\end{aligned}
\end{equation}
where $\text{tr}(.)$ denotes the trace of the matrix.
\subsubsection{Updating regularization parameter $\gamma^2$}
In a similar fashion to noise variance, the regularization parameter $\gamma^2$ follows an inverse-Gamma distribution given by


\begin{equation}
\begin{aligned}
q^k(\gamma^2) = \mathcal{IG}\left(\gamma^2; \frac{N_1}{2} + \eta, \nu + \frac{1}{2} E_{q^k(\bfx)}\left[ \bfx^T \bfDelta^{-1} \bfx \right]\right),
\end{aligned}
\end{equation}
whose mean is given by
\begin{equation}
\begin{aligned}
E_{q^k(\gamma^2)}(\gamma^2) = \frac{\nu + \frac{1}{2}E_{q^k(\bfx)}\left[ \bfx^T \bfDelta^{-1} \bfx \right]}{N_1/2 + \eta - 1}
\end{aligned}
\end{equation}
where 
\begin{equation}
\begin{aligned}
& E_{q^k(\bfx)}\left[ \bfx^T \bfDelta^{-1} \bfx \right] = E_{q^k(\bfx)}(\bfx^T) \bfDelta^{-1} E_{q^k(\bfx)}(\bfx)  \\
& \hspace{3cm} + \text{tr} \left(\bfDelta^{-1} \bfSigma_{q^k(\bfx)}(\bfx)\right).
\end{aligned}
\end{equation}
\subsubsection{Updating the hyperparameter $\beta$}
The hyperparameter $\beta$ follows a Gamma distribution given by

\begin{equation}
\begin{aligned}
q^k(\beta) = \mathcal{G}\left(\beta; \alpha + \alpha_o, \frac{\beta_oE_{q^k(\sigma^2)}(\sigma^2)}{\beta_o + E_{q^k(\sigma^2)}(\sigma^2)}\right),
\end{aligned}
\end{equation}
whose mean is given by
\begin{equation}
\begin{aligned}
E_{q^k(\beta)}(\beta) = \frac{\left(\alpha + \alpha_o\right)\beta_oE_{q^k(\sigma^2)}(\sigma^2)}{\beta_o + E_{q^k(\sigma^2)}(\sigma^2)}.
\end{aligned}
\end{equation}

In Algorithm \ref{alg:VB}, no assumptions were imposed on the posterior approximation of $q(\bfx)$. We can, however, assume as \cite{babacan2009variational, babacan2011variational, molina2006blind, babacan2008parameter, bishop2007blind}, that this distribution is degenerate, i.e., distribution which takes one value with probability one and the rest of the values with probability zero. We can obtain another algorithm under this assumption which is similar to algorithm \ref{alg:VB}.

\begin{algorithm}
\caption{Deconvolution via VB}
\label{alg:VBFinal}
\begin{algorithmic}[1]
\State \textbf{Set} $k = 1$, 
\State \textbf{Initialize} $E_{q^1(\sigma^2)}(\sigma^2), E_{q^1(\gamma^2)}(\gamma^2)$ and $E_{q^1(\beta)}(\beta)$,
\State \textbf{repeat} (k = k + 1)

\State $E_{q^k(\bfx)}(\bfx) = \left(\frac{\bfH^T\bfH}{E_{q^k(\sigma^2)}(\sigma^2)} + \frac{\bfDelta^{-1}}{E_{q^k(\gamma^2)}(\gamma^2)}\right)^{-1} \frac{\bfH^T\bfy}{E_{q^k(\sigma^2)}(\sigma^2)}$
\vspace{0.3cm}
\State $E_{q^k(\sigma^2)}(\sigma^2) = \frac{E_{q^k(\beta)}(\beta) + \norm{\bfy - \bfH E_{q^k(\bfx)}(\bfx)}^2_2}{N_1/2 + \alpha - 1}$
\vspace{0.3cm}
\State $E_{q^k(\gamma^2)}(\gamma^2) = \frac{\nu + \frac{1}{2}\left(E_{q^k(\bfx)}(\bfx)\right)^T \bfDelta^{-1} E_{q^k(\bfx)}(\bfx)}{N_1/2 + \eta - 1}$
\vspace{0.3cm}
\State $E_{q^k(\beta)}(\beta) = \frac{\left(\alpha + \alpha_o\right)\beta_oE_{q^{k+1}(\sigma^2)}(\sigma^2)}{\beta_o + E_{q^{k+1}(\sigma^2)}(\sigma^2)}$
\vspace{0.3cm}
\State \textbf{until} some stopping criterion is satisfied.
\vspace{0.3cm}
\State \textbf{Set}
$\hat{\bfx} = E_{q^k(\bfx)}(\bfx)$, $\hat{\sigma^2} = E_{q^k(\sigma^2)}(\sigma^2)$, $\hat{\gamma^2} = E_{q^k(\gamma^2)}(\gamma^2)$, and $\hat{\beta} = E_{q^k(\beta)}(\beta)$  
\end{algorithmic}
\end{algorithm}

The stopping criterion we use is $\sum_{\mathcal{H} \in \{\bfx, \sigma^2, \beta, \gamma^2\}}^{}\norm{\mathcal{H}^{(k)} - \mathcal{H}^{(k+1)}}_F \leq \epsilon $, where $\epsilon = \sqrt{N_1}\times 10^{-5}$ \cite{afonso2011augmented}.\\

It is clear that using degenerate distribution for $q(\bfx)$ in Algorithm \ref{alg:VBFinal} removes the uncertainty terms of the intensity field estimate. It has been shown that this helps to improve the restoration performance  \cite{babacan2009variational, babacan2011variational, molina2006blind, babacan2008parameter, bishop2007blind}. Moreover, it also reduces the computational complexity as there is no need to compute explicitly the covariance matrix $\bfSigma_{q^k(\bfx)}(\bfx)$ at each iteration. Finally, a few remarks are needed to obtain a fast algorithm. The inverse of the covariance matrix $\bfDelta$ needs to be computed only once before the loop in Algorithm \ref{alg:VBFinal}. We also considered the MATLAB operation $\left(\frac{\bfH^T\bfH}{E^k(\sigma^2)} + \frac{\bfDelta^{-1}}{E^k(\gamma^2)}\right) \backslash (\bfH^T\bfy)$ for the update of the intensity field vector $\bfx$, which is faster than computing the covariance matrix in (\ref{eq:IntensityFieldVB}b), then updating the mean in (\ref{eq:IntensityFieldVB}a). For very big images, diagonal approximation \cite{babacan2011variational} or conjugate gradient \cite{luessi2014variational} can be considered for the update of the intensity field vector $\bfx$.
\vspace*{-0.3cm}
\subsection{ADMM algorithm}
This section describes another alternative to the MCMC algorithm which is based on an optimization algorithm. The latter maximizes the joint posterior distribution \eqref{eq:Posterior_1} $f(\bfOmega|\bfy, \bfphi)$ with respect to (w.r.t.) the parameters of interest, with fixing the hyperparameter vector $\bfphi$, to approximate the MAP estimator of $\bfTheta$, or equivalently, by minimizing the negative log-posterior distribution given by $\mathcal{F} = -\log\left[f (\bfTheta | \bfy\right]$. The resulting optimization problem is tackled using ADMM that sequentially updates the different parameters, which is widely used in the literature for solving imaging inverse problems \cite{afonso2011augmented, figueiredo2010restoration, afonso2010fast}. We rewrite the model as an optimization problem as follows 
\begin{equation} 
\begin{aligned}
& \underset{\bfx}{\text{minimize}}
& & \frac{1}{2}\norm*{\bfH\bfx - \bfy}^2_2 + \lambda \phi(\bfx) + i_{\mathbb{R}+}(\bfx),
\label{eq:ADMM_MCMC_3}
\end{aligned}
\end{equation}
where the regularization function $\phi(\bfx)$ is proportional to the negative logarithm of the intensity field prior considered in \eqref{eq:Bayesian_3} up to an additive constant, i.e. $\phi(\bfx)  = \frac{\bfx^T\bfDelta^{-1}\bfx}{2}$, and $\lambda = \sigma^2 / \gamma^2$ is the regularization parameter. Given this objective function, we write the constrained equivalent formulation as follows

\begin{equation} 
\begin{aligned}
& \underset{\bfu, \bfx}{\text{minimize}}
 \hspace{0.5cm}\frac{1}{2}\norm*{\bfH\bfx - \bfy}^2_2 + \lambda \phi(\bfx) + i_{\mathbb{R}+}(\bfu),\\
& \hspace{0.7cm} \text{subject to}
\hspace{1cm} \bfu = \bfx,
\label{eq:ADMM_MCMC_4}
\end{aligned}
\end{equation}
where $\bfu$ and $\bfx$ are the variables to minimize. In order to solve for $\bfu$ and $\bfx$, we construct the augmented Lagrangian corresponding to \eqref{eq:ADMM_MCMC_4} as follows

\begin{equation} 
\begin{aligned}
\mathcal{L}(\bfu, \bfx, \bfd_1) = \frac{1}{2}\norm*{\bfH\bfx - \bfy}^2_2 + \lambda \phi(\bfx) + i_{\mathbb{R}+}(\bfu) \\
 + \frac{\mu}{2}\norm*{\bfx - \bfu - \bfd_1}^2_2,
\label{eq:ADMM_MCMC_5}
\end{aligned}
\end{equation}
where $\mu > 0$ is a positive parameter. The ADMM algorithm for solving \eqref{eq:ADMM_MCMC_5} is shown in Algorithm (\autoref{alg:ADMM_MCMC_6}). During each step of the iterative algorithm, $\mathcal{L}$ is optimized w.r.t. $\bfu$ (step 3) and $\bfx$ (step 4) and then the Lagrange multipliers are updated (step 6). The stopping criterion we use is $\norm{\bfu^{(k)} - \bfx^{(k)}}_F \leq \epsilon $, where $\epsilon = \sqrt{N_1}\times 10^{-5}$ \cite{afonso2011augmented}.

\begin{algorithm}
\caption{Deconvolution via ADMM}
\label{alg:ADMM_MCMC_6}
\begin{algorithmic}[1]
\State set $k = 0$, choose $\mu > 0, \bfu^{(0)}, \bfx^{(0)}, \text{ and } \bfd^{(0)}_1$
\State \textbf{repeat} (k = k + 1)
\State $\bfu^{(k+1)} = \text{max }{\left(\bfx^{(k)} - \bfd^{(k)}_1, 0\right)}$
\State $\bfx^{(k+1)} = \left(\bfH^T\bfH + \lambda \bfDelta^{-1} + \mu \bfI\right)^{-1}\left[\bfH^T\bfy + \mu\left(\bfu + \bfd^{(k)}_1\right)\right]$
\State $\textbf{Update Lagrange multipliers:}$\\
\hspace{1cm} $\bfd^{(k+1)}_1 = \bfd^{(k)}_1 - \left(\bfx^{(k+1)} - \bfu^{(k+1)}\right)$
\State \textbf{Update iteration} $k \leftarrow k + 1$
\State \textbf{until} some stopping criterion is satisfied.
\end{algorithmic}
\end{algorithm}
\vspace*{-0.3cm}
\section{Non-Linear Interpolation Using Gaussian Process Regression}
\label{sec:GPR}
In order to visually view a meaningful image from the deconvolved intensities, we consider non-linear interpolation based on Gaussian processes (GP) \cite{rasmussen2006gaussian}, since it can provide confidence intervals for each interpolated pixel. A classic choice consists of considering a zero-mean GP with an arbitrary covariance matrix. Here, we choose this covariance matrix to be $\bfDelta^\prime = \bfDelta /\gamma^2$. Precisely, we interpolate using the prior distribution previously defined in \eqref{eq:CovarianceMRFandGP}. If $d_{n, n'}$ is very small, then $\bfDelta^\prime(n, n')$ approaches its maximum $1/ \gamma^2$. If $n$ is distant from $n'$, we have instead $\bfDelta^\prime(n, n') \approx 0$, i.e. the two points are considered to be \emph{a priori} independent. So, for example, during interpolation at new $n_*$ location, distant cores will have negligible effect. The amount of spatial correlation depends on the parameters $\ell$, and $\kappa$, which are estimated in the way we previously mentioned in section \ref{subsubsec:Xprior}.

If we consider $\bfDelta^\prime(\bfz, \bfz) \in \mathbb{R}^{N_1 \times N_1}$, $\bfz=[z_1,\ldots,z_{N_1}]^T$ contains all the positions of all the observed cores (whose estimated intensities are gathered into $\bfx$), and a new spatial location $z_*$ for which we want to predict the intensity $x_*$, the GP can be extended as follows

\begin{equation}
   \begin{bmatrix}
     \bfx \\
     x_*
   \end{bmatrix}
\sim \mathcal{N} \left( \bf0, 
  \begin{bmatrix}
     \bfDelta^\prime(\bfz, \bfz) & \bfDelta^\prime(\bfz, z_*) \\
     \bfDelta^\prime(z_*, \bfz) & 1/\gamma^2
  \end{bmatrix}\right),
  \label{eq:GPR_Cond}
\end{equation}
where $\bfDelta^\prime(\bfz, z_*)= \bfDelta^\prime(z_*, \bfz)^T \in \mathbb{R}^{N_1}$. Eq. \eqref{eq:GPR_Cond} shows that the conditional  distribution of each predicted intensity given the previously estimated intensities, follows a Gaussian distribution $x_*|\bfx \sim \mathcal{N} \left(\bfmu, \bfSigma\right)$ whose mean and variance are given by

\begin{equation}
\begin{aligned}
\bfmu & = & \bfDelta^\prime(z_*, \bfz)\bfDelta^\prime(\bfz, \bfz)^{-1}\bfx, \\
\bfSigma & = & 1/\gamma^2 - \bfDelta^\prime(z_*, \bfz)\bfDelta^\prime(\bfz, \bfz)^{-1}\bfDelta^\prime(\bfz, z_*).
\label{eq:GPR_MVAR}
\end{aligned}
\end{equation}

By setting $\bfx = \hat{\bfx}$, the mean in \eqref{eq:GPR_MVAR} is finally used to estimate each interpolated intensity, while the variance is used to provide additional information (measure of uncertainty) about the interpolated intensity values.

The Matlab implementations of this paper are provided at \url{https://sites.google.com/site/akeldaly/publications}.

\section{Simulations Using Synthetic Data}
\label{sec:Results}
\subsection{Data creation}
\begin{figure}[!h]
	\centering
		\includegraphics[width=0.8\textwidth]{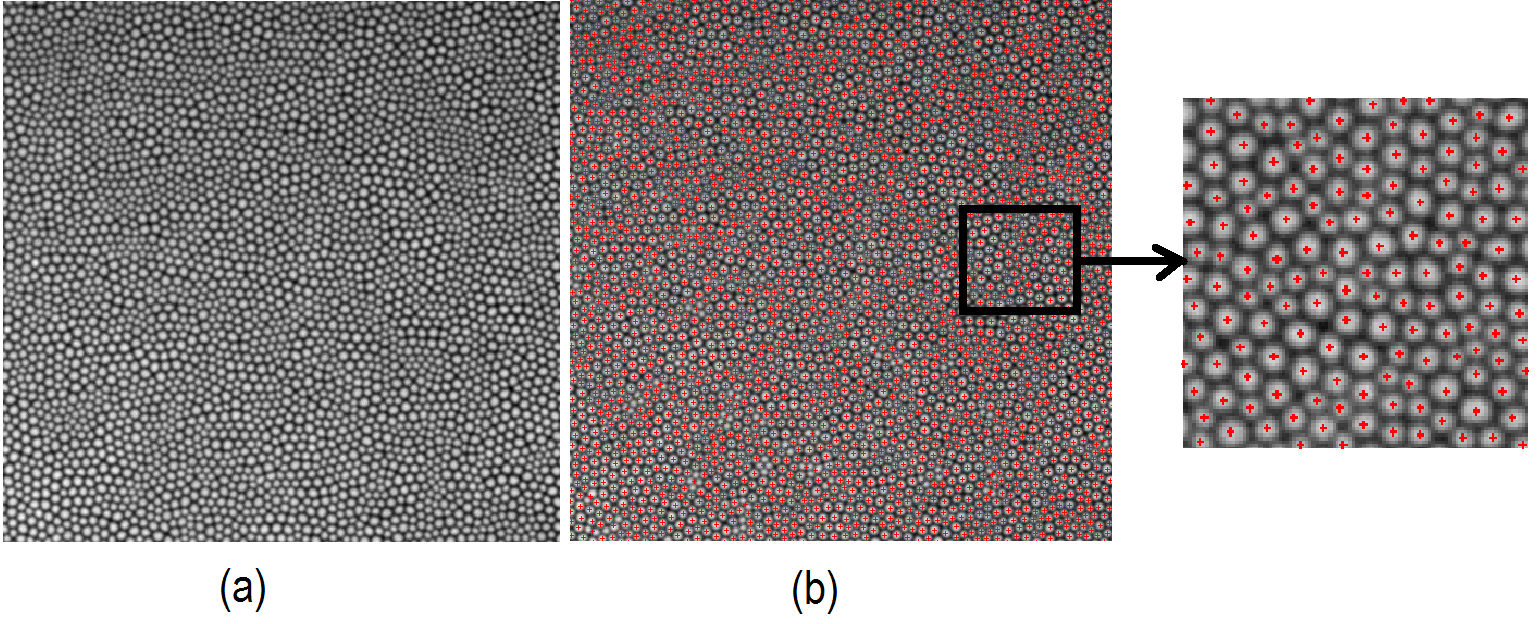}
	\caption{(a) Example of $512 \times 512$ pixels image of the endomicroscopy system (b) Image with detected fiber core centres superimposed (red crosses).}
	\label{fig:Sampling_Pattern}
\end{figure}

The performance of the proposed methods is investigated by reconstructing a standard test image. A subsampled version of this image is obtained by considering the sampling pattern of an actual endomicroscopy system, as illustrated in Fig. \ref{fig:Sampling_Pattern}. This figure provides an example of a homogeneous region imaged through  Alveoflex (Mauna Kea Technologies, France) fiber bundle \cite{le2004towards}\cite{ayache2006processing}. Such image is used for calibration and to identify the number and positions of the fiber cores. The build-in MATLAB function ``vision.BlobAnalysis'' was used to detect central fibre core pixels.

\begin{figure}[h]
\centering  
\subfloat []{\label{subfig:Original}\includegraphics[scale = 0.35]{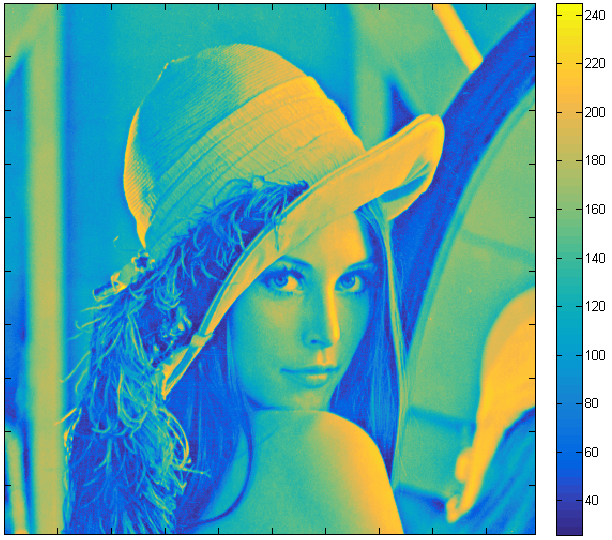}}
\subfloat []{\label{subfig:CHX}\includegraphics[scale = 0.35]{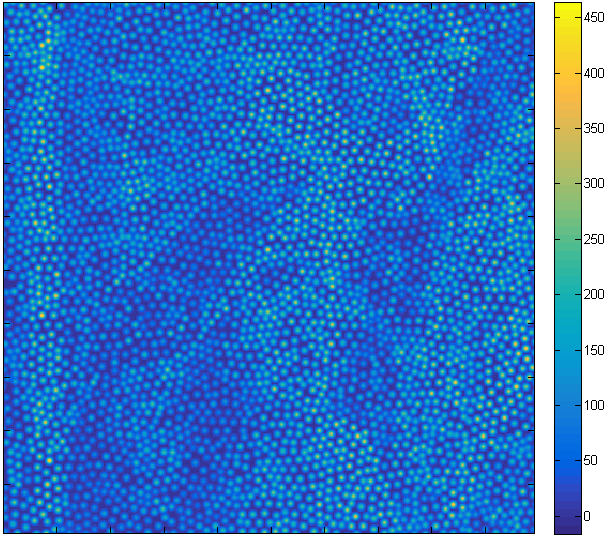}}
\caption{Creation of the synthetic data: (a) Original image (b) example of final system output with $\sigma^2_\bfH = 20$ and $\sigma^2_N = 10$.}
\label{fig:PreProcessing}
\end{figure}

Fig. \ref{fig:PreProcessing} shows the original Lena image (left) and an example of system output (right) after applying the model in Eq. \eqref{eq:Formulation_1}. This image is formed by creating a binary mask in which a value of 1 is assigned to pixels corresponding to the central pixels of each core in Fig. \ref{fig:Sampling_Pattern}(b), and zero otherwise. This mask is then multiplied point by point by the Lena image in Fig. \ref{fig:PreProcessing}(a) in order to obtain the subsampled image. The model in Eq. \eqref{eq:Formulation_1} is then applied to obtain an image that simulates the system's output which is shown in Fig. \ref{fig:PreProcessing}(b). This image is created using subsampled intensities corresponding to 1.29\% of the original Lena image. For simulated data, we considered a Gaussian spatial blurring kernel with one size $\sigma^2_\bfC = 2$ in all the simulations.

\subsection{Performance analysis}
The performance discriminator adopted in this work to measure the quality of the deconvolved fiber cores is the root mean square error (RMSE), which is computed using intensities at the core locations using
\begin{equation}
\begin{aligned}
\text{RMSE}(\bfx, \hat{\bfx}) = \sqrt{\frac{\sum_{n = 1}^{N_1} \left(\bfx(n) - \hat{\bfx}(n)\right)^2}{N_1}},
\end{aligned}
\end{equation}
where $\bfx$ and $\hat{\bfx}$ are vectors of the subsampled reference Lena image and its deconvolved version respectively, and $N_1$ is the number of fibre cores.

For synthetic data, in order to check the performance of the algorithm with different cross coupling effects, different values of $\alpha_\bfH$ and $\beta_\bfH$ in \eqref{eq:CrossCoupling_Form} can be considered. However, this can be simplified by considering a 2D Gaussian kernel defined by \eqref{eq:CrossCoupling_Form1}
\begin{equation}
\begin{aligned}
[\bfH]_{i,j} = \exp\left(\frac{-d_{i,j}^2}{2\sigma^2_\bfH}\right),
\label{eq:CrossCoupling_Form1}
\end{aligned}
\end{equation}
since it involves only one variable to change, namely $\sigma^2_\bfH$ (representing a squared distance, in pixels). This is equivalent to setting $\beta_\bfH = 2 \text{ and } \alpha^2_\bfH = \alpha^2_\bfH/2$. Note that this simplification is considered only for synthetic data in order to assess the influence of the kernel width. The generalized Gaussian cross coupling kernel $\bfH$ defined in \eqref{eq:CrossCoupling_Form} will be considered for real data.

\begin{figure}
\centering   
\subfloat []{\label{subfig:Reconstructed1}\includegraphics[scale=0.45]{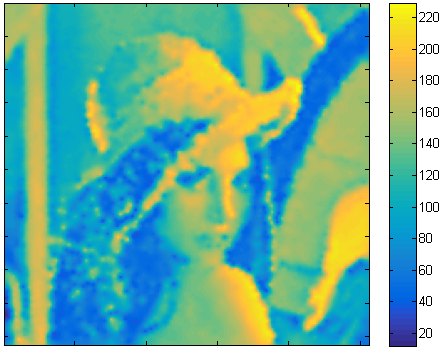}}
\subfloat []{\label{subfig:Reconstructed2}\includegraphics[scale=0.45]{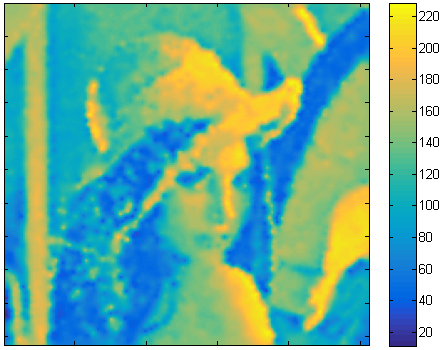}}
\\
\subfloat []{\label{subfig:ConfidenceIntervals1}\includegraphics[scale=0.45]{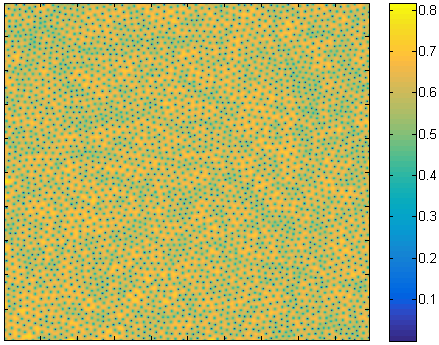}}
\subfloat []{\label{subfig:ConfidenceIntervals2}\includegraphics[scale=0.45]{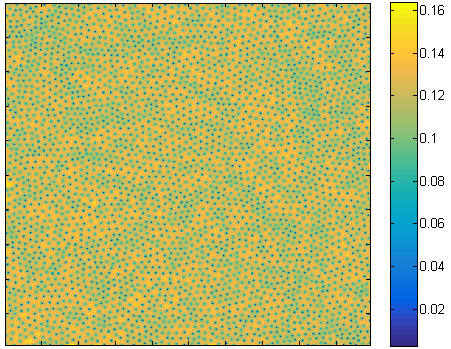}}
\caption{Examples of interpolated samples by GP after deconvolution (a) $\sigma^2_N = 0 \text{ and } \sigma^2_H = 1$, and (b) $\sigma^2_N = 10\text{ and }\sigma^2_H = 20$, and the corresponding confidence interval images.}
\label{fig:AlgorithmsOutputSynthetic}
\end{figure}

The three methods showed similar results in terms of RMSE and interpolated images. The following shows the VB method's results. Fig. \ref{fig:AlgorithmsOutputSynthetic} shows examples of interpolated intensities after deconvolution using GP in the noise-free case ($\sigma^2_{N} = 0$) and noisy case ($\sigma^2_{N} = 10$) and different values of $\sigma^2_\bfH$, with the corresponding confidence interval images. we can observe that the structure of the Lena image can be recovered in the two cases. Moreover, in the confidence interval images, we can observe that as we go away from central cores, the confidence interval of the interpolated intensities decreases.

\begin{figure}
	\centering
		\includegraphics[width=0.6\textwidth]{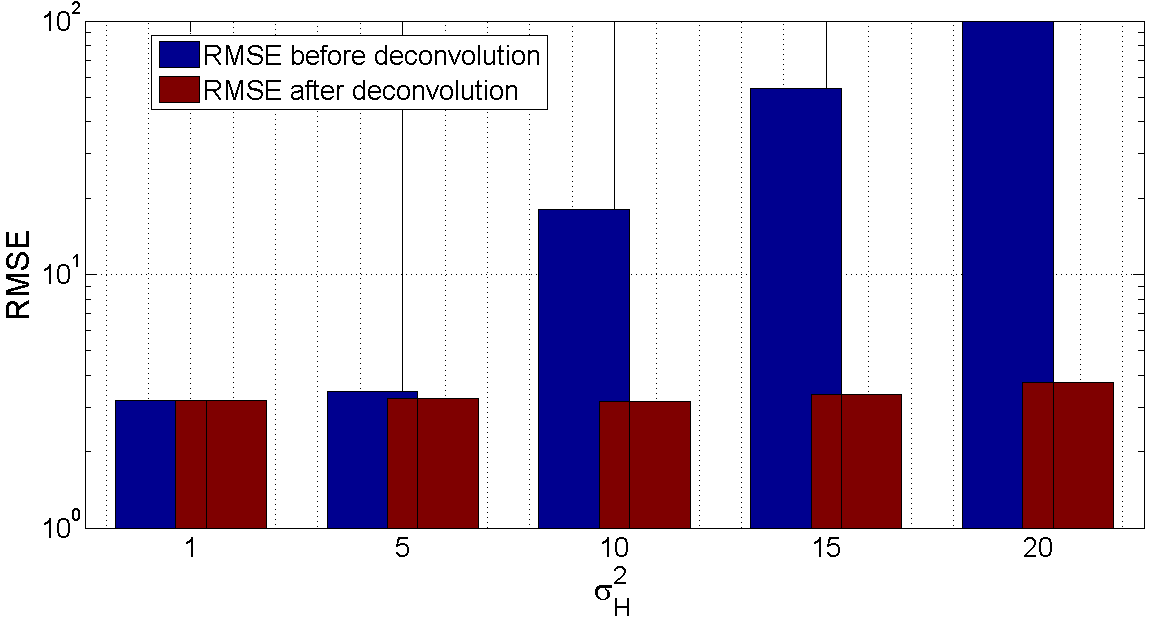}
	\caption{Plot of RMSEs before and after deconvolution (in-log scale) versus $\sigma^2_\bfH$ at $\sigma^2_{N} = 10$.}
	\label{fig:b4_and_after_deconv_Fibrecores}
\end{figure}

In order to measure the performance of the algorithms, we consider different noise variances ($\sigma_N^2$) as well as different cross coupling effects ($\sigma_\bfH^2$). Fig. \ref{fig:b4_and_after_deconv_Fibrecores} shows the RMSE (in log-scale) before and after deconvolution versus $\sigma^2_\bfH$ at $\sigma^2_{N} = 10$. We can observe that all of the methods are very effective since the RMSE after deconvolution is always lower than that before deconvolution. Moreover, the gain increases with cross coupling.

\begin{figure}
\centering     
\subfloat []{\label{subfig:Fixed_H_ADMM_HX_Fibrecores}\includegraphics[scale=0.35]{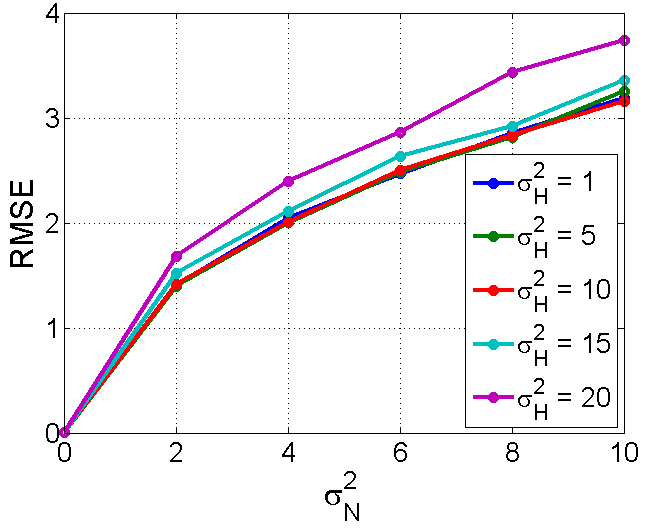}}
\subfloat []{\label{subfig:Fixed_N_ADMM_HX_Fibrecores}\includegraphics[scale=0.35]{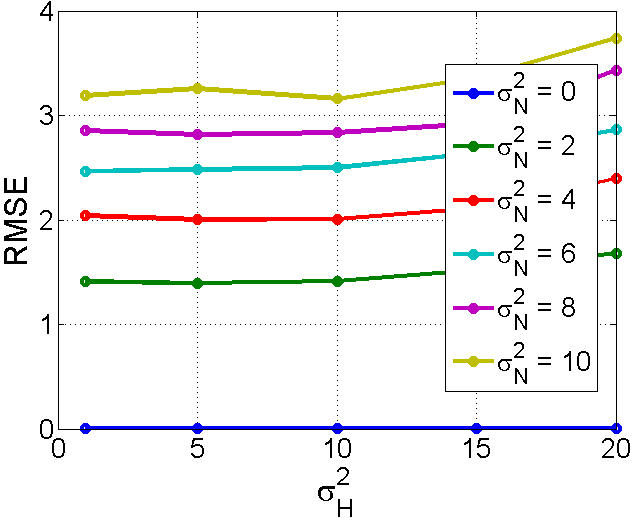}}
\caption{Plot of RMSEs after deconvolution (a) versus $\sigma^2_{N}$ at fixed $\sigma^2_{H}$, and (b) versus $\sigma^2_{H}$ at fixed $\sigma^2_{N}$.}
\label{fig:ADMM_Fibercores}
\end{figure}

In order to analyze the effect of noise variance and cross coupling separately, we fix one of them and change the other as shown in Fig. \ref{fig:ADMM_Fibercores}. In this figure, we show plots of RMSEs after deconvolution for different $\sigma^2_N$ at fixed $\sigma^2_\bfH$ and vice versa. In Fig. \ref{fig:ADMM_Fibercores}(a), we can observe that there is roughly a linear relationship between RMSE and $\sigma^2_N$ at fixed $\sigma^2_\bfH$. Moreover, the behaviour at $\sigma^2_\bfH = 1, 5, 10 \text{ and } 15$ is almost the same. In Fig. \ref{fig:ADMM_Fibercores}(b), we can observe that RMSE is fairly constant as $\sigma^2_\bfH$ increases at constant $\sigma^2_N$. Furthermore, it starts to increase as $\sigma^2_N$ increases but still remains constant when changing $\sigma^2_\bfH$.

For the MCMC method, in all of the simulations in this paper including the real datasets, $N_{\text{MC}} = 1500$, including $N_{\text{bi}} = 500$, which were determined visually from preliminary runs, were used. For the ADMM method, different regularization parameter values are tested, we pick up the one corresponding to the lowest RMSE.
\vspace*{-0.3cm}
\subsection{Comparison}
In this section, we compare the three proposed methods for deconvolution and restoration of OEM images. The comparison is conducted in terms of RMSE before and after deconvolution, as well as in terms of computation time.

\begin{figure}
	\centering
		\includegraphics[scale=0.4]{{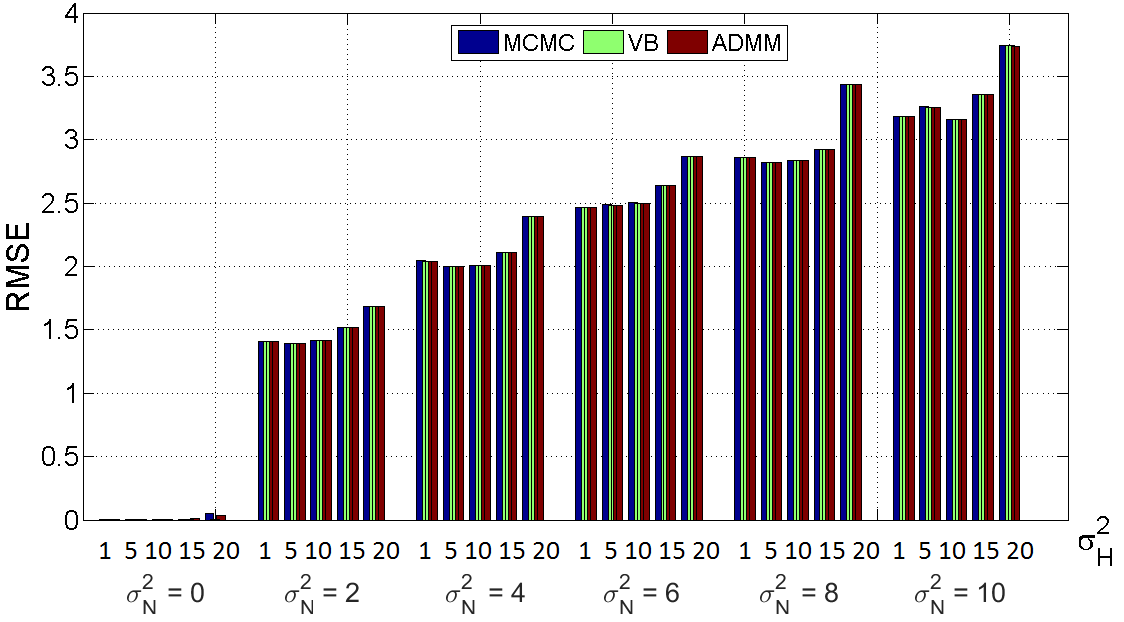}}
	\caption{Plot of RMSEs before and after deconvolution for the three methods versus $\sigma^2_{N}$ as well as $\sigma^2_{H}$.}
	\label{fig:Comparison}
\end{figure}

Fig. \ref{fig:Comparison} compares RMSEs after deconvolution versus different $\sigma^2_N$ as well as different $\sigma^2_\bfH$. We can observe that for all of the methods, as $\sigma^2_N$ increases at constant $\sigma^2_\bfH$, RMSE increases. On the other hand, at fixed $\sigma^2_N$, RMSE seems to be roughly constant for $\sigma^2_\bfH = 1, 5, \text{ and } 10$, then, it starts to increase as $\sigma^2_\bfH$ increases. It is clear that all the methods behave similarly in terms of RMSE.

\autoref{tab:Comparison_Time} shows the average computation time (in seconds) of the three proposed methods. The experiments were conducted on ACER core-i3-2.0 GHz processor laptop with 8 GB RAM. It is clear that the MCMC method is the most computationally expensive method. The ADMM method is second, and the VB the least. Despite the relatively high computation time of the MCMC method, it is a parameter free method compared to the ADMM-based method in which the regularization parameter $\lambda$ should be chosen carefully. The VB approach is considered to be the best compared to MCMC and ADMM, it can provide similar RMSE but with lower computation complexity, moreover, it is fully automatic in the sense that it can estimate the hyperparameters associated with the parameters as mentioned previously in section \ref{subsec:VB}. 

\begin{table}[h]
\centering
\caption{The average computation time (in seconds) of the three proposed methods. In order to maintain a fair comparison between the three algorithms, the computational time of the ADMM algorithm corresponds to the duration of five runs (used to select the best regularization parameter among the five values).}
\begin{tabular}{|c|c|c|c|}
\hline
\textbf{Method} & MCMC & ADMM & VB \\ \hline
\textbf{\begin{tabular}[c]{@{}c@{}}Computation time (sec.)\end{tabular}} & 3100 & 35.51 & \textbf{5.12} \\ \hline
\end{tabular}
\label{tab:Comparison_Time}
\end{table}

Although the MCMC and ADMM algorithms can estimate the noise variance and model hyperparameters, in practice these
parameters are very difficult to estimate accurately, (specifically $\sigma^2$ and $\gamma^2$) due to the similarity between $\bfH^T\bfH$ and $\bfDelta^{-1}$ in (\ref{eq:IntensityPrior}b) and (\ref{eq:IntensityFieldVB}b). Therefore, we have to make an informed choice about one of these parameters, specifically the choice of the hyperparameters $\alpha$, $\alpha_0$ and $\beta_0$ in \eqref{eq:Bayesian_6} and \eqref{eq:Bayesian_5}. In Fig. \ref{fig:ADMM_Fibercores}(b), we observe that the RMSEs in practise are close to the true noise standard deviation, and hence the noise variance can be inferred.

\subsection{Robustness}
To test the robustness of the proposed methods, we create the data using a specific $\sigma^2_H$ and we deconvolve using different values. Following this strategy, we create the data using $\sigma^2_H = 10$ and we deconvolve using $\sigma^2_H = 6, 8, 10, 12, \text{ and } 14$. The three estimation approaches showed similar results.

\begin{figure}
\centering     
\subfloat []{\label{subfig:Robustness_FixedH}\includegraphics[scale=0.3]{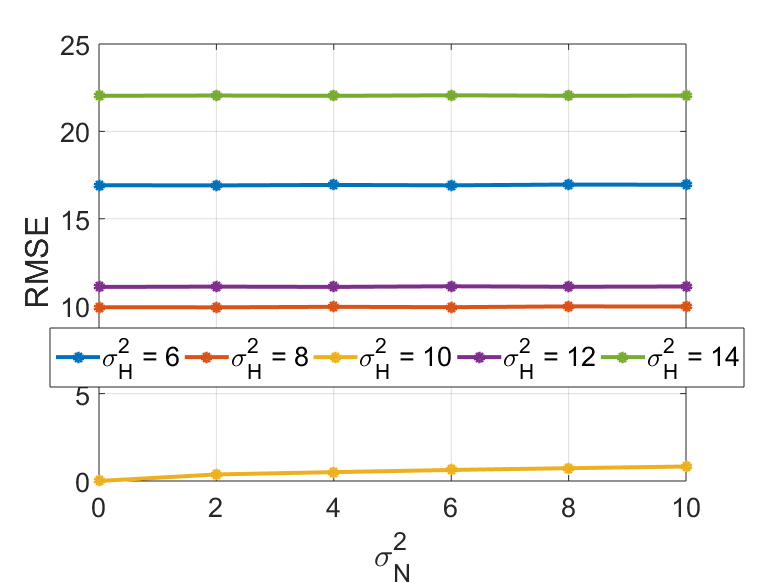}}
\subfloat []{\label{subfig:Robustness_FixedN}\includegraphics[scale=0.3]{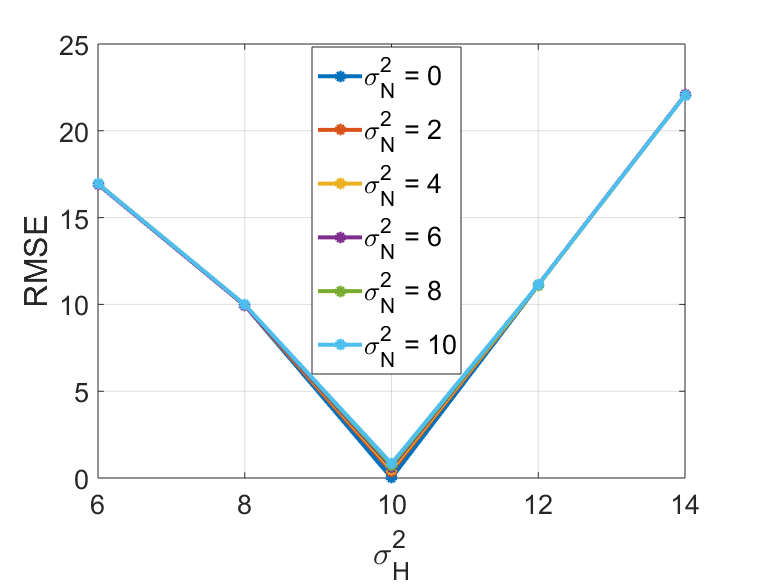}}
\caption{Plots of RMSEs between the central fiber cores in the original Lena image and the deconvolved central fiber cores versus (a) $\sigma^2_N$ at fixed $\sigma^2_H$, and (b) $\sigma^2_H$ at fixed $\sigma^2_N$ of the IIo method.}
\label{fig:Robustness}
\end{figure}

Fig. \ref{fig:Robustness} shows plots of RMSE after deconvolution versus $\sigma^2_N$ at fixed $\sigma^2_\bfH$ and vice versa. In Fig. \ref{fig:Robustness}(a), we can observe that the noise variance has no effect on the deconvolution in the tested interval as RMSE is constant at fixed $\sigma^2_\bfH$. In Fig. \ref{fig:Robustness}(b), there is an approximately linear relationship between RMSE and $\sigma^2_\bfH$ at constant $\sigma^2_N$. Furthermore, lower values of $\sigma^2_H$ than the one we created the data with (i.e., $\sigma^2_\bfH = 6 \text{ and } 8$) yield lower RMSE than higher ones (i.e., $\sigma^2_\bfH = 12 \text{ and } 14$). In other words, it is slightly better to underestimate $\sigma^2_\bfH$ than to overestimate it.

We observe that deconvolution using the value we created the data with ($\sigma^2_\bfH = 10$) yields the minimum RMSE. Moreover, RMSE after deconvolution is always lower than that before deconvolution except for $\sigma^2_\bfH = 14$ at which it is higher.

\vspace*{-0.2cm}
\section{Simulations Using Real Data}
\label{sec:RealData}
The performance of the proposed methods has been evaluated on two real datasets; the 1951 USAF resolution test chart and \emph{ex vivo} human lung tissue. Both of them were collected using OEM system \cite{krstajic2016two} with monochrome detection (Grasshopper3 camera GS3-U3-23S6M-C, Point Grey Research, Canada) and 470 nm LED illumination (M470L3, Thorlabs Ltd, UK) for lung autofluorescence excitation. Excised human lung tissue was placed in a well plate. Human tissue was used with regional ethics committee (REC: 13/ES/0126) approval and was retrieved from the periphery of specimens taken from lung cancer resections. In order to adjust the cross coupling kernel parameters $\alpha_\bfH \text{ and } \beta_\bfH$, a study was performed to measure, analyze and quantify inter-core coupling within coherent fibre bundles \cite{perperidis2017characterization}. This study showed how light is spread over the neighbouring cores, and gave statistical analysis on coupling percent in neighbouring cores. It showed that around 61\% of transmitted light remains in the central core, around 34\% in the first neighbouring cores, around 4\% in the second neighbouring cores, and less than 1\% in the third, fourth and fifth neighbouring cores. This leads to fixing $\alpha_\bfH = 4$ (in pixels) and $\beta_\bfH = 0.8$.

\subsection{1951 USAF resolution test chart}
\begin{figure}[!h]
\centering     
\subfloat []{\label{subfig:USAF_Original}\includegraphics [width=4.2cm,height=3.52cm]{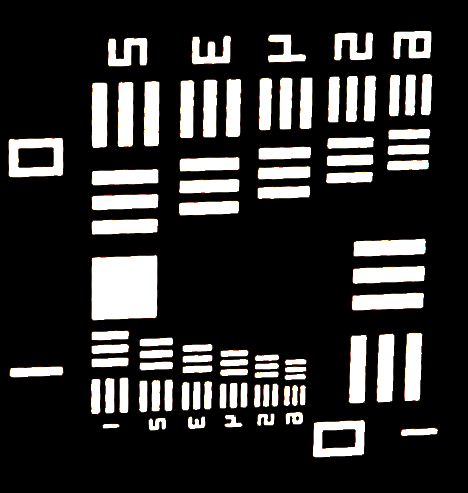}}
\subfloat []{\label{subfig:Original_Proteus}\includegraphics[scale=0.18]{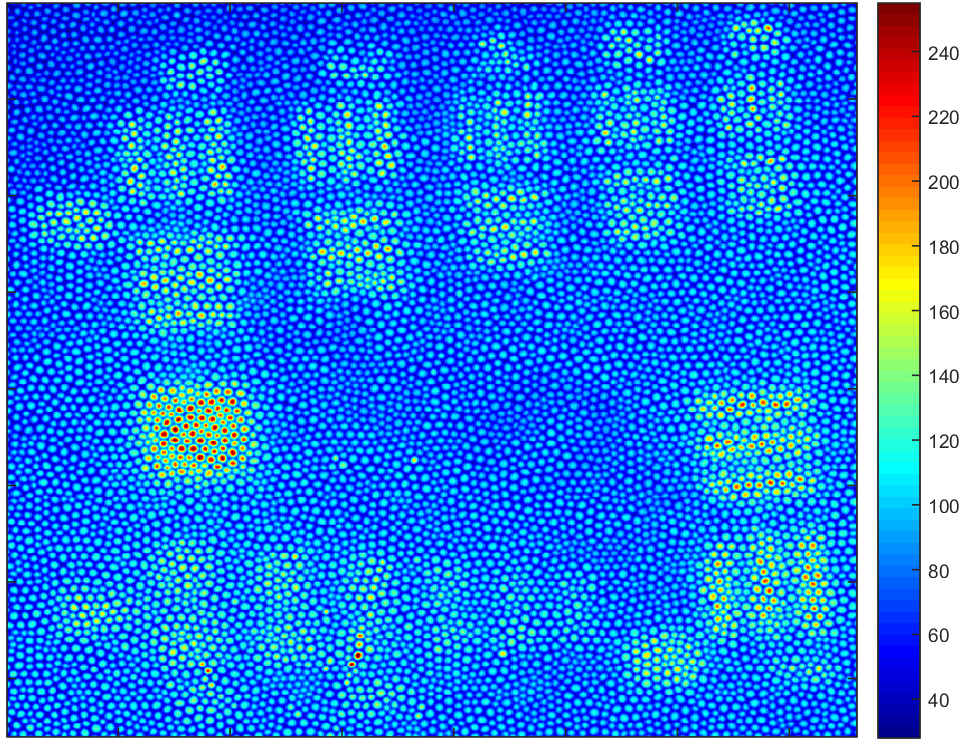}}
\caption{(a) Scanned image of an USAF 1951 Resolution test chart. (b) The 1951 USAF resolution test chart imaged by the OEM system.}
\label{fig:test}
\end{figure}

The 1951 USAF chart is a resolution test pattern set by US Air Force in 1951. It is widely accepted to test the resolution of optical imaging systems such as microscopes, cameras and image scanners \cite{SilverFast}. Fig. \ref{fig:test} (a) shows the original USAF resolution test chart used in the project. The resulting image obtained by fiber bundle is shown in Fig. \ref{fig:test} (b) with image size $760 \times 760$ and is composed of 7,776 fiber cores ($1.34\%$ of the image).

\begin{figure}[!h]
\centering     
\subfloat []{\label{subfig:test1}\includegraphics[width=4.1cm,height=3.3cm]{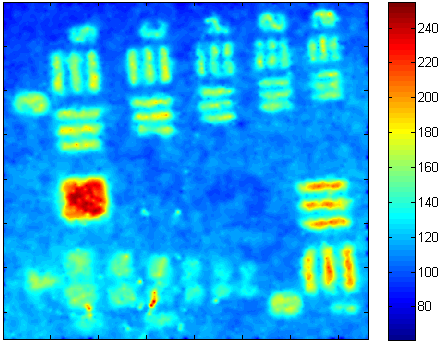}}
\subfloat []{\label{subfig:Interpolated_Proteus}\includegraphics[scale=0.365]{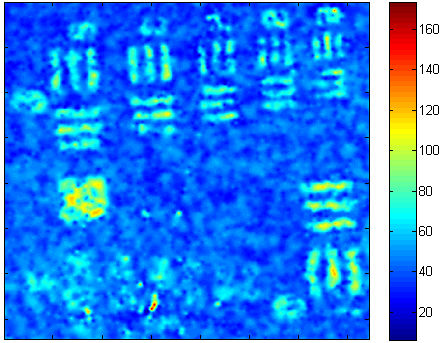}}\\
\subfloat []{\label{subfig:ConfIntervB4}\includegraphics[width=4.1cm,height=3.3cm]{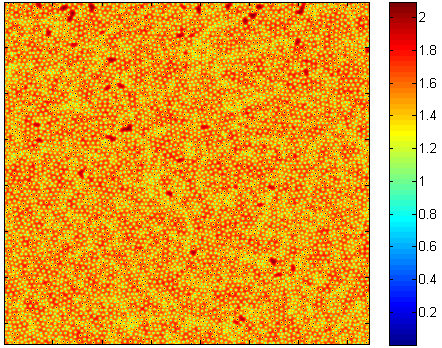}}
\subfloat []{\label{subfig:ConfIntervAfter}\includegraphics[scale=0.365]{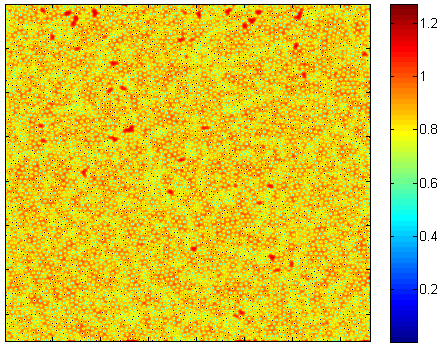}}
\caption{Non-linear interpolation (a) before, and (b) after deconvolution, and their corresponding confidence intervals in (c), and (d) respectively.}
\label{fig:USAF_Microendoscopy}
\end{figure}

A non-linear interpolation based on GP of central core intensities of the image in Fig. \ref{fig:test}(b) is presented in Fig.\ref{fig:USAF_Microendoscopy}(a), with the corresponding confidence intervals image in Fig.\ref{fig:USAF_Microendoscopy}(c). We can observe the blurring which is caused by the cross coupling effect as well as the sparsity of the data.

The outputs of the MCMC, VB, and ADMM algorithms are very similar. Thus, we show the results of the VB method. Fig. \ref{fig:USAF_Microendoscopy}(b) shows an example of one of the output images with the corresponding confidence intervals in Fig. \ref{fig:USAF_Microendoscopy}(d). The set of ticker strips (top left corner of the image) is now better resolved and the overlap between them is reduced. The small set of strips which is at the bottom could not be resolved, which gives an indication about the resolving resolution of this endomicroscopy system. Regions of high uncertainty (which appear as blobs in dark red) are where there may be no cores or they are dead, this in addition to the irregular core sampling are the reasons for some strips appear a bit fragmented.
\vspace*{-0.4cm}
\subsection{Ex vivo human lung tissues}
\begin{figure}[!h]
\centering   
\subfloat []{\label{subfig:Pulmonary_Original}\includegraphics[width=4.5cm,height=3.45cm]{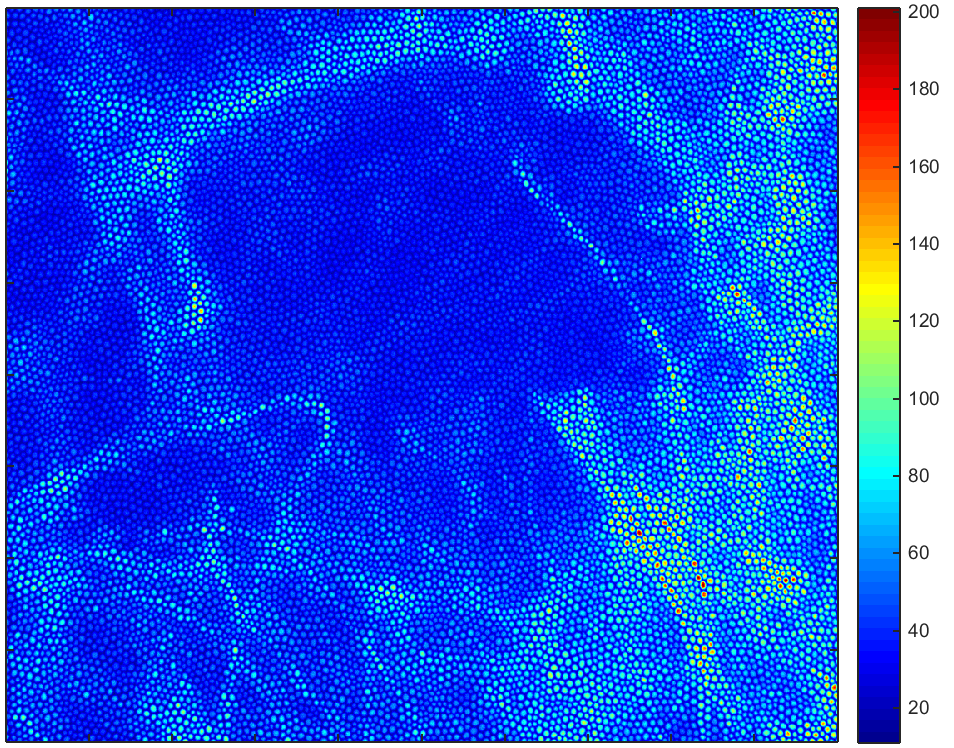}}  
\subfloat []{\label{subfig:GPUSAFBefore}\includegraphics[width=4.5cm,height=3.45cm]{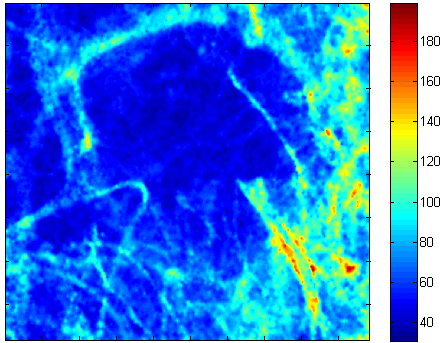}}\\
\subfloat []{\label{subfig:GPUSAFAfter}\includegraphics[scale=0.38]{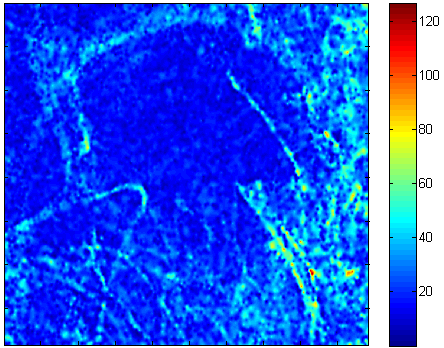}}
\subfloat []{\label{subfig:GPUSAFBeforeConf}\includegraphics[width=4.6cm,height=3.45cm]{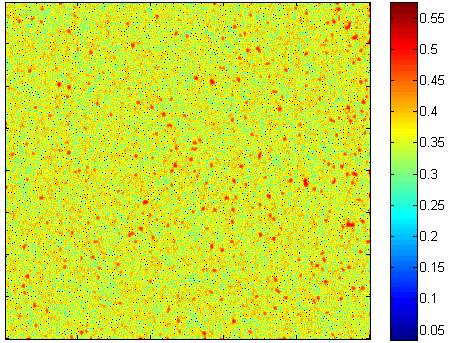}}
\caption{(a) \emph{Ex vivo} lung tissue imaged by the endomicroscopy system \cite{krstajic2016two}. Non-linear interpolation (b) before, and (c) after deconvolution, (d) the confidence intervals of the image in (c).}
\label{fig:Pulmonary_Microendoscopy}
\end{figure}

Fig. \ref{fig:Pulmonary_Microendoscopy}(a) shows the output image of the OEM system. Image size is $1000 \times 800$ and is composed of 13,343 fiber cores ($1.66\%$ of the image). Non-linear interpolation based on GP of central core intensities is presented in Fig.\ref{fig:Pulmonary_Microendoscopy}(b). Similar to the USAF resolution test chart, we aim at reducing cross coupling effect as well as getting a more resolved image.

Similar to the USAF resolution test chart results, the outputs of the MCMC, VB, and ADMM algorithms are very similar. We only show the results of the VB method. Fig. \ref{fig:Pulmonary_Microendoscopy}(c) shows an example of interpolated deconvolved samples using GP. The lung structure is now better resolved and more sharper than before deconvolution. Moreover, confidence intervals are shown in Fig. \ref{fig:Pulmonary_Microendoscopy}(d). We can observe that as we move away from the central cores, the confidence of the interpolated intensities decreases and vice versa.

Table \ref{tab:ComputationTimeRealData} provides the computation time of the 1951 USAF resolution test chart and the \emph{ex vivo} lung tissue image. It is clear that the VB is still the fastest despite the change of the images size.

\begin{table}[!h]
\centering
\caption{Computation time (in seconds) for the real data. In order to keep a fair comparison between the three algorithms, the computational times of the ADMM algorithm correspond to the duration of five runs (used to select the best regularization parameter among five values).}
\begin{tabular}{|c|c|c|c|}
\hline
\textbf{Dataset/Method} & MCMC & ADMM & VB \\ \hline
USAF chart & $1.12\times 10^5$ & $250$ & $\textbf{5.9}$ \\ \hline
Lung tissue & $1.46\times 10^6$ & $870$ & \textbf{16.05} \\ \hline
\end{tabular}
\label{tab:ComputationTimeRealData}
\end{table}

\section{Conclusion and Future Work}
\label{sec:Conclusion}
This paper introduced a hierarchical Bayesian model and three estimation algorithms for the deconvolution of optical endomicroscopy images. The deconvolution accounts and compensates for fibre core cross coupling which causes major image degradation in this type of imaging. The resulting joint posterior distribution was used to approximate the Bayesian estimators. First, a Markov chain Monte Carlo procedure based on a Gibbs sampler algorithm was used to sample the posterior distribution of interest and to approximate the MMSE estimators of the unknown parameters using the generated samples. Second, a variational Bayes approach to approximate the joint posterior distribution by minimizing the Kullback-Leibler divergence was used. Third, an approach based on an alternating direction method of multipliers was used to approximate the maximum a posteriori estimators. The three algorithms showed similar estimation performance while providing different characteristics, the MCMC and VB based approaches are fully automatic in the sense that they can jointly estimate the hyperparameters associated with the priors, however, the MCMC based approach showed high computational complexity which could be overcome by the VB and ADMM approaches. Although the ADMM approach has low computational complexity, it is semi-supervised in the sense that the hyperparameters associated with the priors need to be chosen carefully by the user. A non-linear interpolation approach based on Gaussian processes was considered to restore the full images from the samples to provide a meaningful image for interpretation. In the future, we will consider temporal information while deconvolving. Accounting for the different core sizes is also clearly an interesting route currently under investigation. 

\vspace*{-0.3cm}
\section*{Acknowledgement}
This work was supported in parts by the Engineering and Physical Sciences Research Council (EPSRC, United Kingdom) Interdisciplinary Research Collaboration grant EP/K03197X/1 and by the Royal Academy of Engineering under the Research Fellowship scheme (RF201617/16/31). We would like to thank the reviewers for their helpful comments that helped in improving the quality of the manuscript.

\ifCLASSOPTIONcaptionsoff
  \newpage
\fi

\bibliographystyle{ieeetran}
\bibliography{references}

\end{document}